\documentclass[journal]{IEEEtran}
\usepackage{}
\usepackage{lineno,hyperref}
\usepackage{graphicx}
\usepackage{epstopdf}
\usepackage{bm}
\usepackage{multirow}
\usepackage{threeparttable}
\usepackage[subfigure]{tocloft}
\usepackage{subfigure}
\ifCLASSINFOpdf
\else
\fi
\begin{document}

\title{Robust Facial Landmark Detection by Multi-order Multi-constraint Deep Networks}


\author{Jun~Wan, Zhihui~Lai, Jing~Li, Jie~Zhou, Can~Gao
	
	
	\thanks{This work is supported by the National Natural Science Foundation of China(Grant No. 62002233, 62076164, 61802267, 61976145 and 61806127),  the Shenzhen Municipal Science and Technology Innovation Council (Grant No. JCYJ20180305124834854), the Natural Science Foundation of Guangdong Province (Grant No. 2019A1515111121, 2017A030313367, 2018A030310451 and 2018A030310450) and the China Postdoctoral Science Fundation (Grant No. 2020M672802). Corresponding author: Zhihui Lai.}
	\thanks{J. Wan is with the College of Computer Science and Software Engineering, Shen zhen University, Shenzhen, 518060, China, and the School of Mathematics and Statistics, Hanshan Normal University, Chaozhou, 521041, China.(e-mail: junwan2014@whu.edu.cn).}
	\thanks{Z. Lai, J. Zhou and C. Gao are with the College of Computer Science and Software Engineering, Shen zhen University, Shenzhen, 518060, China, and the Shenzhen Institute of Artificial Intelligence and Robotics for Society, Shenzhen, 518060, China.(e-mail: lai\_zhi\_hui@163.com, jie\_jpu@163.com, 2005gaocan@163.com)}
	\thanks{J. Li is with the School of Computer Science, Wuhan University, Wuhan, 430072, China (e-mail: leejingcn@whu.edu.cn).}
}

\markboth{Journal of \LaTeX\ Class Files,~Vol.~14, No.~8, August~2015}%
{Shell \MakeLowercase{\textit{et al.}}: Bare Demo of IEEEtran.cls for IEEE Journals}

\maketitle
\begin{abstract}
Recently, heatmap regression has been widely explored in facial landmark detection and obtained remarkable performance. However, most of the existing heatmap regression-based facial landmark detection methods neglect to explore the high-order feature correlations, which is very important to learn more representative features and enhance shape constraints. Moreover, no explicit global shape constraints have been added to the final predicted landmarks, which leads to a reduction in accuracy. To address these issues, in this paper, we propose a Multi-order Multi-constraint Deep Network (MMDN) for more powerful feature correlations and shape constraints learning. Specifically, an Implicit Multi-order Correlating Geometry-aware (IMCG) model is proposed to introduce the multi-order spatial correlations and multi-order channel correlations for more discriminative representations. Furthermore, an Explicit Probability-based Boundary-adaptive Regression (EPBR) method is developed to enhance the global shape constraints and further search the semantically consistent landmarks in the predicted boundary for robust facial landmark detection. It's interesting to show that the proposed MMDN can generate more accurate boundary-adaptive landmark heatmaps and effectively enhance shape constraints to the predicted landmarks for faces with large pose variations and heavy occlusions. Experimental results on challenging benchmark datasets demonstrate the superiority of our MMDN over state-of-the-art facial landmark detection methods. The code has been publicly available at \url{https://github.com/junwan2014/MMDN-master}.
\end{abstract}

\begin{IEEEkeywords}
Heatmap regression, feature correlations, shape constraints, heavy occlusions, boundary-adaptive regression.
\end{IEEEkeywords}

\IEEEpeerreviewmaketitle

\section{Introduction}

\IEEEPARstart{F}{acial} landmark detection (FLD), also known as face alignment, refers to locating the predefined landmarks (eye corners, nose tip, mouth corners, etc.) of a face. As a topical issue in computer vision, FLD has drawn much attention recently as it can provide rich geometric information for other face analysis tasks such as face recognition \cite{huang2017learning, cao2019fr}, face verification \cite{Sun2014DeepLF}, face frontalization \cite{Hassner2015EffectiveFF}  and 3D face reconstruction \cite{Garrido2016ReconstructionOP}. 
\begin{figure}[t]
\begin{center}
\includegraphics[width=0.98\linewidth]{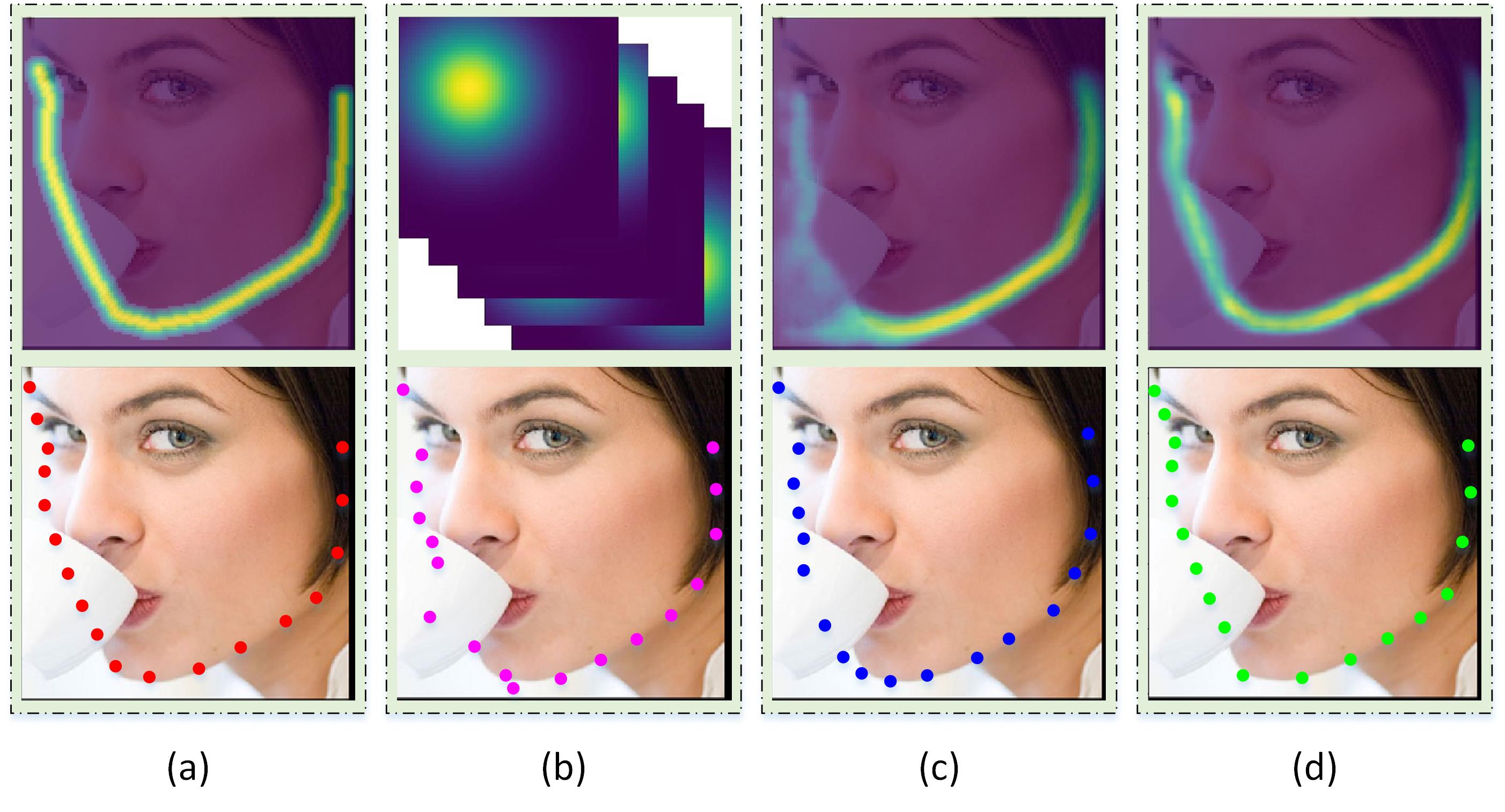}
\end{center}
\vspace{-1em}
   \caption{Comparison of predicted landmarks in face contour of (a) ground truth, (b) heatmap regression, (c) boundary regression and (d) our method. The heatmap regression-based method and the boundary regression-based method with mean squared error fail to capture shape constraints among landmarks. With the proposed MMDN, our method can generate more accurate boundary heatmaps and effectively enhance the shape constraints, thus achieving more accurate facial landmark detection.}
\label{fig1}
\vspace{-1em}
\end{figure}
\\\indent In recent years, heatmap regression-based methods \cite{Yang2017StackedHN,Dong2018StyleAN,Liu2019SemanticAF} have become one of the mainstream approaches to solve FLD problems. Due to the effective encoding of part constraints and context information, heatmap regression-based FLD methods have achieved considerable performance on faces under variations on poses, expressions, occlusions and illuminations. However, as shown in Fig. \ref{fig1}, for faces with severe occlusions or extremely large poses, the accuracy of heatmap regression-based FLD methods is greatly reduced as 1) this kind of method reduces or even lose shape/geometric constraints between landmarks due to predicting landmarks in isolation. 2) occlusions or large poses may mislead neural networks on feature representation learning and shape/geometric constraints learning. Recent works \cite{Gao2018GlobalSP, Dai2019SecondOrderAN} have shown the second-order statistics (information) in deep convolutional neural networks can effectively model the spatial and channel correlations between features and mine the useful information contained in the features themselves. They can fully explore more representative information and are more discriminative than first-order statistics. However, how to introduce the high-order information for enhancing shape/geometric constraints (generating more effective landmark and boundary heatmaps) and improving the performance of facial landmark detection are still open questions.
\\\indent In this paper, we propose a Multi-order Multi-constraint Deep Network (MMDN) for both implicit and explicit shape constraints learning (see Fig. \ref{fig2}). Specifically, we propose an Implicit Multi-order Correlating Geometry-aware (IMCG) model to enhance shape constraints by introducing the multi-order spatial correlations and multi-order channel correlations. The multi-order spatial correlations contain both the auto-correlation of intra-layer features and the cross-correlation of inter-layer features, which can be used to capture hierarchical patterns and attain global receptive fields. The multi-order channel correlations can be utilized to selectively emphasize informative features and suppress less useful ones by considering both the first-order and second-order statistics of features, thus performing feature recalibration and improving the quality of representations. Moreover, an Explicit Probability-based Boundary-adaptive Regression (EPBR) method is developed to handle facial landmark detection by integrating boundary constraints and specific semantically consistent constraints into the final predicted landmarks, enhancing the robustness to faces with large poses and heavy occlusions. Therefore, our method obtains better robustness and accuracy compared with other state-of-the-art FLD methods.
\\\indent The main contributions of this work are summarized as follows:
\\\indent 1) With the well-designed IMCG model, multi-order spatial and channel correlations can be introduced to implicitly enhance geometric constraints and context information for more effective feature representations when dealing with complicated cases, especially for faces with extremely large poses and heavy occlusions.
\\\indent 2) An EPBR method is presented to explore how to incorporate the boundary heatmap regression with landmark heatmap regression to generate boundary-adaptive landmark heatmaps through the proposed searching mechanism, and further explicitly add both boundary and semantically consistent constraints to the final predicted landmarks. Hence, the EPBR method is able to enhance the robustness when the dataset is corrupted by strong noise.
\\\indent 3) A novel algorithm called MMDN is developed to seamlessly integrate the IMCG model and EPBR method into a Multi-order Multi-constraint Deep Network to handle facial landmark detection in the wild. With more effective representations and regression, our algorithm outperforms state-of-the-art methods on challenging benchmark datasets such as COFW \cite{Burgosartizzu2013Robust}, 300W \cite{Sagonas2016300FI}, AFLW \cite{Zhu2016UnconstrainedFA} and WFLW \cite{Wu2018LookAB}.
\\\indent The rest of the paper is organized as follows. Section \textbf{II} gives an overview of the related work. Section \textbf{III} shows the proposed method, including the IMCG model and the EPBR method. A series of experiments are conducted to evaluate the performance of the proposed method in Section \textbf{IV}. Finally, Section \textbf{V} concludes the paper.
\begin{figure*}[!t]
  \centering
  \includegraphics[width=16cm,height=7.2cm]{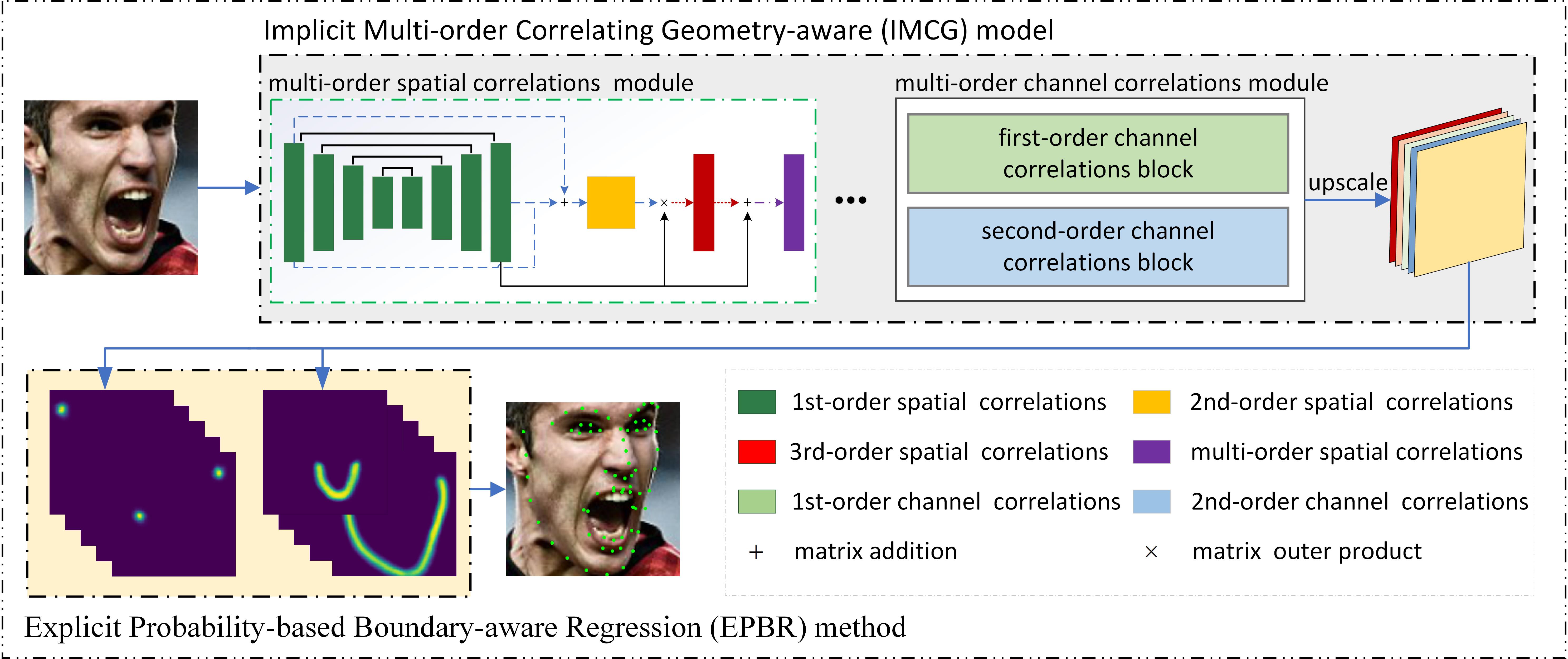}
  \centering
  \caption{ The overall architecture of our proposed MMDN. The IMCG model can implicitly enhance geometric constraints and context information for more effective feature representations by fusing the multi-order spatial correlations and the multi-order channel correlations. The EPBR method can generate more accurate and robust landmark heatmaps and boundary heatmaps by incorporating the Jensen-Shannon divergence loss with the searching mechanism. Then, by integrating the IMCG model and EPBR method into a Multi-order Multi-constraint Deep Network via a seamless formulation, our MMDN can outperform state-of-the-art methods.}
  \label{fig2}
  \vspace{-1em}
\end{figure*}
\section{Related Work}
During the past decades, plenty of facial landmark detection (FLD) methods have been proposed in the computer vision area. In general, existing methods can be categorized into three groups: model-based methods, coordinate regression-based methods, and heatmap regression-based methods.
\\\indent\textbf{Model-based methods.} The model-based FLD methods represented by active shape model \cite{cootes1995active}, active appearance model \cite{cootes2001active} and constrained local model \cite{Cristinacce2006FeatureDA} depend on the goodness of the error function. They use parametric models (facial shape model \cite{cootes1995active}, facial appearance model \cite{cootes2001active} or part model \cite{Cristinacce2006FeatureDA}) to constrain the shape variations and improve the performance of FLD. Active shape model \cite{cootes2001active} uses Principal Component Analysis to establish a facial shape model and an appearance model, then these two models are combined to enhance the shape constraints and texture information for improving the performance of facial landmark detection. Constrained local model \cite{Cristinacce2006FeatureDA} firstly constructs a shape model and a patch model, then it can detect more accurate landmarks by searching and matching them on the neighborhood regions around each landmark in the initial shape. By jointly optimizing a global shape model and a part-based appearance model with an efficient and robust Gaussian Newton optimization, the performance of Gauss-Newton deformable part models \cite{Tzimiropoulos2014GaussNewtonDP} can be improved and the computation costs have been reduced. However, these methods are sensitive to faces under large poses and partial occlusions. 
\\\indent\textbf{Coordinate regression-based methods.} This category of FLD method directly learns the mapping from facial appearance features to the landmark coordinate vectors by using different kinds of regressors such as fern \cite{cao2014face}, random forest \cite{ren2014face}, convolution neural networks \cite{Merget2018RobustFL, shi2018deep}, recurrent neural networks \cite{Trigeorgis2016MnemonicDM} or hourglass network \cite{Yang2017StackedHN}. In explicit shape regression \cite{cao2014face} and local binary features \cite{ren2014face}, fern and random forest are used separately to extract features and regress. With these two excellent regressors, they both achieve good results. In tasks-constrained deep convolutional network \cite{zhang2014facial}, the convolution neural network is used for feature extraction, and the author proposes to utilize auxiliary attributes in the fitting process and applies the multi-task learning framework to improve the performance of face alignment. In mnemonic descent method \cite{Trigeorgis2016MnemonicDM}, a convolutional recurrent neural network is used to extract the task-based features and enhance the generalization ability of the algorithm against faces with large poses and partial occlusions. Merget et al. \cite{Merget2018RobustFL} directly introduce global context into a fully convolutional neural network, and then the kernel convolution and the dilated convolutions are utilized to smooth the gradients and reduce over-fitting, thus enhancing its robustness. Wu et al. \cite{Wu2018LookAB} firstly introduce an adversarial network and message passing layers to generate more robust boundary heatmaps. Then the generated boundary heatmaps will be used as part of the input to enhance shape constraints and further improve the accuracy of FLD. In occlusion-adaptive deep networks \cite{Zhu2019RobustFL}, the geometry-aware module, the distillation module and the low-rank learning module are combined to overcome the occlusion problem for robust FLD. With these discriminative features and effective constraints, these methods are more robust to faces with variations on poses, expressions, illuminations and partial occlusions.
\\\indent\textbf{Heatmap regression-based methods.} Heatmap regression-based FLD methods \cite{Yang2017StackedHN, Dong2018StyleAN, Liu2019SemanticAF} directly predict the landmark coordinates by traversing the generated landmark heatmaps. This kind of method can better encode the part constraints and context information, and effectively drive the network to focus on the important part in facial landmark detection, thus achieving state-of-the-art accuracy. Yang et al. \cite{Yang2017StackedHN} firstly adopt a supervised face transformation to reduce the variance of the regression target and then use a stacked hourglass network to increase the capacity of the regression model. By paying more attention to the feature with high confidence in an explicit manner, the robustness of this method is enhanced. Dong et al. propose a style aggregated network \cite{Dong2018StyleAN} that can generate different styles of images from one image. The generated images and the original ones are used together to train a more robust model that can explicitly handle problems caused by the variation of image styles in FLD. Liu et al. \cite{Liu2019SemanticAF} propose a novel latent variable optimization strategy to find the semantically consistent annotations and alleviate random noises during the training stage, the ground-truth shape can be updated and the predicted shape becomes more accurate.
\\\indent Until now, most heatmap regression-based FLD methods do not fully utilize the inter-feature dependencies and inter-channel dependencies, and they ignore the high-order information higher than the first-order information, thus they cannot effectively explore the discriminative power of features. Recent works \cite{Gao2018GlobalSP, Dai2019SecondOrderAN} have shown the second-order statistics in deep convolutional neural networks are more discriminative than the first-order ones. Moreover, heatmap regression-based FLD methods predict landmarks in isolation, which causes some predicted landmarks to deviate from the facial boundary, and the shape constraints among landmarks may be lost during the training. Inspired by the above observations, we proposed a Multi-order Multi-constraint Deep Network by incorporating the IMCG model with the EPBR method for robust FLD.
\section{Robust Facial Landmark Detection by Multi-order Multi-constraint Deep Networks}
In this section, we first elaborate on the proposed Implicit Multi-order Correlating Geometry-aware (IMCG) model in \textbf{ Section III. A}, and then present the Explicit Probability-based Boundary-adaptive Regression (EPBR) method in \textbf{ Section III. B}. \textbf{Section III. C} illustrates the proposed Multi-order Multi-constraint Deep Network. Finally, \textbf{Section III. D} optimizes the proposed MMDN.
\begin{figure}[t]
	\begin{center}
		\includegraphics[width=0.96\linewidth]{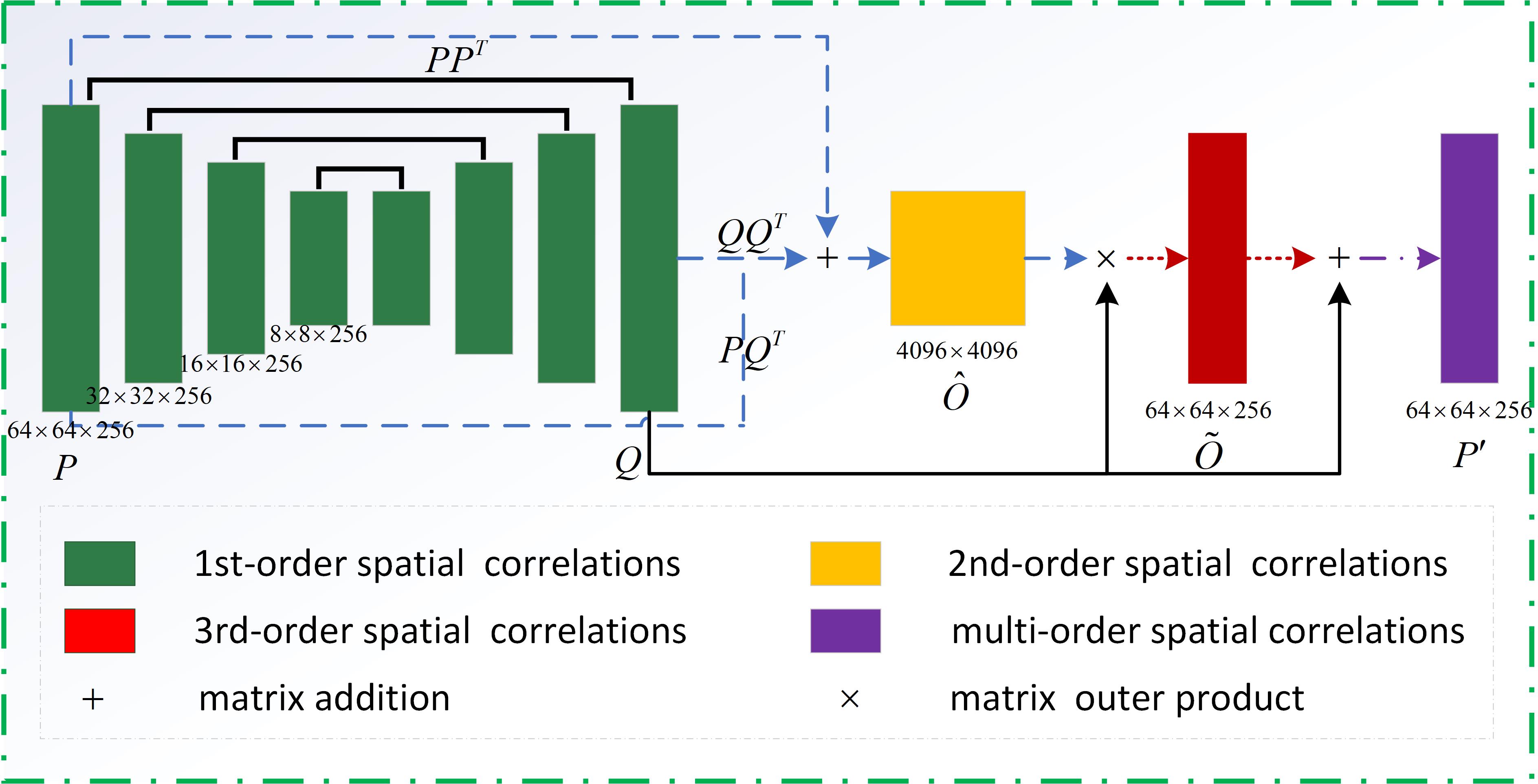}
	\end{center}
	\caption{The network structure of the multi-order spatial correlations module. By introducing the multi-order spatial correlations, our proposed multi-order spatial correlations module can be used to capture hierarchical patterns and attain global receptive fields while preserving local details, thus better encoding the geometric constraints and context information for robust facial landmark detection.}
	\label{fig3}
	\vspace{-1em}
\end{figure}
\subsection{Implicit Multi-correlating Geometry-aware (IMCG) model}
\begin{figure*}[!t]
	\centering
	\includegraphics[width=0.9\linewidth]{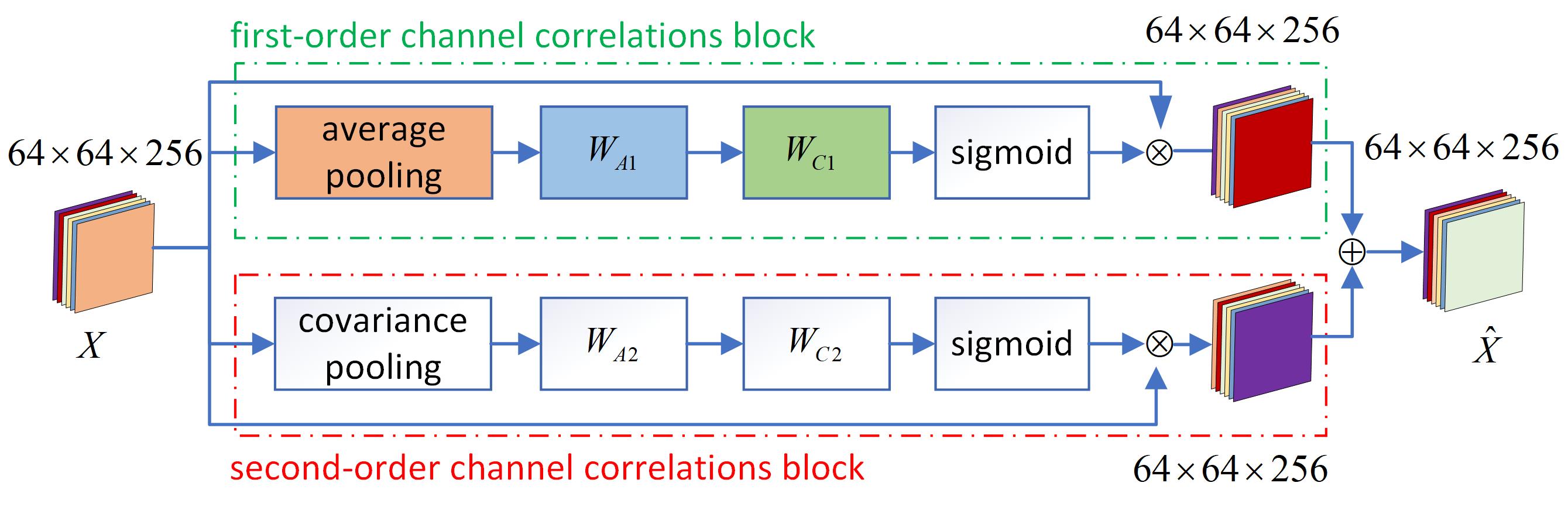}
	\centering
	\vspace{-1em}
	\caption{The network structure of the multi-order channel correlations module. The multi-order channel correlations module mainly contains a first-order channel correlations block and a second-order channel correlations block. These two blocks can better model the inter-channel dependencies by introducing the first-order and second-order channel correlations with the global average and covariance pooling operations. Then, with the fusion of the first-order and second-order channel correlations, the multi-order channel correlations can selectively emphasize informative channels and suppress less useful ones, i.e., perform effective channel re-calibration. }
	\label{fig4}
	\vspace{-1em}
\end{figure*}

Heatmap regression-based methods \cite{Yang2017StackedHN, Dong2018StyleAN, Liu2019SemanticAF} have achieved state-of-the-art accuracy as they can effectively encode the part constraints and context information. However, they suffer from performance degradation under faces with extremely large poses and heavy occlusions, as in these cases the extracted features are not robust enough and the facial geometric constraints (e.g., part constraints and global constraints) among landmarks may be lost. More recently, the high-order information \cite{Gao2018GlobalSP, Wang2019DeepGG} has been shown to be beneficial to obtain more robust and discriminative representations for many computer vision tasks. However, how to effectively introduce and utilize the high-order information to enhance shape constraints for robust FLD is still challenging. Hence, in this paper, a Multi-order Correlating Geometry-aware (IMCG) model is proposed to explore more discriminative representations and more effective geometric constraints by introducing the high-order information, i.e., the multi-order spatial correlations and the multi-order channel correlations. The proposed IMCG model contains multiple multi-order spatial correlations modules (see Fig. \ref{fig3}) and one multi-order channel correlations module (see Fig. \ref{fig4}), which can be integrated into a stacked hourglass network to generate more effective and accurate heatmaps for robust FLD.

\subsubsection{Multi-order Spatial Correlations Module}
The stacked hourglass network \cite{Yang2017StackedHN, Wu2018LookAB} is able to extract multi-scale discriminative features for its bottom-up and top-down structure and can function as a regressor to predict the final landmarks. The proposed multi-order spatial correlations module follows an hourglass network unit (see Fig. \ref{fig3}). We use $P$ and $Q$ to separately represent the input and output of the hourglass network unit, and $P, Q \in {R^{N \times D}}$. $N = W \times H$, where $W$, $H$ and $D$ represent the width, height and channels of the feature maps, respectively. $P$ and $Q$ represent the first-order spatial correlations that can be introduced by max or average pooling of features in neural networks. ${P_i}$ denotes the $i$-th row of the matrix $P$ and represents the feature corresponding to the $i$-th pixel in the feature maps. ${P_i}{P_j}^T$ can be regarded as the pairwise interaction of two features in the feature maps. Therefore, we use the outer product of $P$ and $P^T$ to represent the auto-correlation of intra-layer features for capturing geometric correlations of facial landmark regions. $P{Q^T}$ is used to denote the cross-correlation of inter-layer features for capturing landmark geometric correlations at different scales (deep layer and shallow layer). As the outer product operation is similar to a quadratic kernel expansion and is indeed a non-local operation, it can be used to effectively model local pairwise feature correlations for capturing long-range dependencies \cite{ Zhu2019RobustFL, lin2017bilinear}. Therefore, $P{P^T}$, $P{Q^T}$ and $Q{Q^T}$ can effectively preserve long-range geometric constraints and we compute the sum of them to construct the second-order spatial correlations which can be expressed as follows:
\begin{small}
\begin{equation}
\hat O = softmax(P{P^T} + P{Q^T} + Q{Q^T})
\end{equation}
\end{small}where $\hat O \in {R^{N \times N}}$ represents the second-order spatial correlations. The second-order spatial correlations can capture the auto-correlation of intra-layer features and the cross-correlation of inter-layer features, which helps effectively mine the information inherent in different convolutional layers and explore more discriminative representations for enhancing geometrical constraints.
\\\indent To better explore the spatial correlations, we multiply $\hat O$ and $Q$ to obtain the third-order spatial correlations $\tilde O$. The third-order spatial correlations can be regarded as the high-order cross-layer attention feature maps, i.e., transforming the pairwise feature correlations $\hat O$ to the high-order cross-layer attention that can be acted on the original output feature maps $Q$. Based on this, we derive the multi-order spatial correlation $P'$ which contains the first-order, second-order and third-order spatial correlations and can be expressed as follows:
\begin{small}
	\begin{equation}
	\tilde O = \hat OQ
	\end{equation}
	\vspace{-1em}
\end{small}
\begin{small}
\begin{equation}
P' =  \lambda \tilde O + Q = \lambda \hat OQ + Q 
\end{equation}
\end{small}where $P'$ denotes the multi-order spatial correlations. $\lambda$ is initialized as 0 and a larger weight is gradually learned to assign to $\lambda$ during training. Hence, by fusing the first-order, second-order and third-order spatial correlations,  the multi-order spatial correlations module can capture effective hierarchical patterns, which would be further used to enhance geometric constraints. Then, the multi-order spatial correlations module is integrated into stacked hourglass networks, so that the multi-order spatial correlations can be updated and propagated to obtain more effective geometric constraints for robust FLD.
\subsubsection{Multi-order Channel Correlations Module}
With multiple multi-order spatial correlations modules, we can obtain effective spatial correlations and robust representations for enhancing the geometric constraints. Then, to fully explore the discriminative abilities of features, we further propose a multi-order channel correlations module to adaptively re-calibrate channel-wise feature maps by explicitly modeling inter-channel dependencies. As shown in Fig .\ref{fig4}, the multi-order channel correlations module mainly contains the first-order channel correlations block and the second-order channel correlations block. By introducing the global average pooling and global covariance pooling \cite{Li2017TowardsFT} operations, the first-order channel correlations and the second-order channel correlations can be introduced to effectively model the inter-channel dependencies. Then, with the further fusion of the first-order and second-order channel correlations, the multi-order channel correlations module can selectively emphasize informative features and suppress less useful ones, i.e., perform more effective channel re-calibration.
\\\indent\textbf{Global covariance pooling.} Compared to average pooling, covariance pooling \cite{Li2017TowardsFT} can better explore the distributions of features and realize the full potential of CNN. Moreover, covariance pooling can introduce the second-order statistics which are more discriminative and representative than the first-order statistics. Therefore, we begin by briefly reviewing the global covariance pooling operation. Let $X$ denote the output (with ReLU) of the IMCG model, where $X \in N \times D$ and $N = W \times H$. $W$, $H$ and $D$ are used to denote the width, height and channels of the feature maps, respectively. Then the sample covariance matrix can be computed as:
\begin{small}
\begin{equation}
\Sigma  = {X^T}\bar IX
\end{equation}
	\vspace{-1em}
\end{small}
\begin{small}
\begin{equation}
\bar I = \frac{1}{N}\left( {I - \frac{1}{N}\bm{1}} \right)
\end{equation}
\end{small}where $I$ and $\bm{1}$ are the $N \times N$ identity matrix and matrix of all ones, respectively. $\Sigma $ is a symmetric positive semi-definite matrix, which has a unique square root and can be computed accurately by Eigenvalue Decomposition or Singular Value Decomposition. The whole process is illustrated as follows:
\begin{small}
\begin{equation}
\Sigma  = U\Lambda {U^T}
\end{equation}
\vspace{-1em}
\end{small}
\\where $U$ is an orthogonal matrix and $\Lambda  = diag\left( {{\gamma _i}} \right)$. $\Lambda $ denotes the diagonal matrix of eigenvalues ${\gamma _i}$ of $\Sigma $. Then the square root of $\Sigma $ can be computed as follows:
\begin{small}
\begin{equation}
Y = U{\Lambda ^{{1 \mathord{\left/{\vphantom {1 2}} \right.\kern-\nulldelimiterspace} 2}}}{U^T}
\end{equation}
\end{small}where ${Y^2}{\rm{ = }}\Sigma $. To obtain $Y$, we need to perform Eigenvalue Decomposition or Singular Value Decomposition on $\Sigma $. However, both Eigenvalue Decomposition or Singular Value Decomposition are not well supported on GPU, which leads to inefficient training. Therefore, according to \cite{Li2017TowardsFT}, Newton-Schulz Iteration \cite{Higham2008FunctionsOM} is applied to achieve a fast computing of the square root $Y$. Specifically, given ${Y_0} = \Sigma $ and ${Z_0} = I$, for $k = 1 \cdots K$, the Newton-Schulz iteration is then updated alternately as follows:
\begin{small}
\begin{equation}
{Y_k} = \frac{1}{2}{Y_{k - 1}}\left( {3I - {Z_{k - 1}}{Y_{k - 1}}} \right)
\end{equation}
\vspace{-1em}
\end{small}
\begin{small}
\begin{equation}
{Z_k} = \frac{1}{2}\left( {3I - {Z_{k - 1}}{Y_{k - 1}}} \right){Z_{k - 1}}
\end{equation}
\end{small}where $k$ is the number of iterations. After several iterations, ${Y_k}$ and ${Z_k}$ can converge to ${Y_{k-1}}$ and ${Z_{k-1}}$, respectively. However, the Newton-Schulz iteration only converges locally. According to \cite{Li2017TowardsFT}, we first pre-normalize $\Sigma$ by trace or Frobenius norm, and then perform the post-compensation procedure to compensate the data magnitude caused by pre-normalization, thus producing the final normalized covariance matrix
\begin{small}
\begin{equation}
\hat Y = \sqrt {tr\left( \Sigma  \right)} {Y_K}
\end{equation}
\end{small}where $tr\left( \Sigma  \right)$ denotes the trace of $\Sigma$. With such global covariance pooling, our proposed multi-order channel correlations module is ready to utilize the second-order statistics to enhance the representative ability of features.
\\\indent The proposed multi-order channel correlations module contains a first-order channel correlations block and a second-order channel correlations block (see Fig. \ref{fig4}). The normalized covariance matrix $\hat Y$ characterizes the second-order channel correlations, which can be used as a channel descriptor by the multi-order channel correlations module to perform channel re-calibration. At the same time, the first-order channel correlations can be introduced by the first-order channel correlations block. Then, by integrating the first-order channel correlations block and the second-order channel correlations block into a multi-order channel correlations module, the multi-order channel correlations can be utilized to perform effective channel re-calibration and enhance the representative capability of features.
\\\indent \textbf{First-order channel correlations  block.} Let $X = \left[ {{x_1},{x_2}, \cdots , {x_d} ,\cdots,{x_D}} \right]$, the first-order channel correlations ${\kappa ^{st}} \in {R^{D \times 1}}$ can be obtained by performing global average pooling to $X$ (see Fig. \ref{fig4})). Then the $d$-th dimension of ${\kappa ^{st}}$ can be computed as:
\begin{small}
\begin{equation}
\kappa _d^{st} = {Pool_{gap}}\left( {{x_d}} \right) = \frac{1}{{H \times W}}\sum\limits_{w = 1}^W {\sum\limits_{h = 1}^H {{x_d}\left( {w,h} \right)} }
\end{equation}
\end{small}where ${Pool_{gap}}$ denotes the global average pooling, and the first-order channel correlations can be introduced by the ${Pool_{gap}}$ operation. After that, a simple gating mechanism \cite{Hu2017SqueezeandExcitationN} with a sigmoid activation is employed to learn a non-mutually-exclusive channel relationship, which ensures that multiple channels are allowed to be emphasized (rather than enforcing a one-hot activation). The whole process can be stated as:
\begin{small}
\begin{equation}
{s^{st}} = \sigma \left( {{W_{A1}}ReLU\left( {{W_{C1}}{\kappa ^{st}}} \right)} \right)
\end{equation}
\end{small}where $\sigma $ denotes the sigmoid activation and $s^{st}$ denotes the scaling factor. ${W_{A1}}$ and ${W_{C1}}$ are the parameters of convolution layers, whose channel dimensions are set to ${D \mathord{\left/ {\vphantom {D 4}} \right. \kern-\nulldelimiterspace} 4}$ and $D$, respectively. The output of the first-order channel correlations block can be stated as follows:
\begin{small}
\begin{equation}
X_d^{st} = s_d^{st}{x_d}
\end{equation}
\vspace{-1em}
\end{small}
\\where $s_d^{st}$ and ${x_d}$ denote the scaling factor and the feature map in the $d$-th channel. With the first-order channel correlations block, the first-order channel correlations are able to perform channel recalibration.
\\\indent\textbf{Second-order channel correlations block.} Let $\hat Y = \left[ {{y_1},{y_2}, \cdots ,{y_d}, \cdots ,{y_D}} \right]$, ${y_d}$ can be used to represent the second-order channel correlations among the $d$-th channel and all channels. The $d$-th dimension of the second-order channel correlations $\kappa _d^{nd}$ is computed as:
\begin{small}
\begin{equation}
\kappa _d^{nd} = \frac{1}{D}\sum\nolimits_d {{y_d}}
\end{equation}
\vspace{-1em}
\end{small}
\\\indent To fully exploit feature inter-dependencies from the aggregated information by global covariance pooling, we apply a gating mechanism. As explored in \cite{Hu2017SqueezeandExcitationN}, the simple sigmoid function can serve as a proper gating function, which can be illustrated as:
\begin{small}
\begin{equation}
{s^{nd}} = \sigma \left( {{W_{A2}}ReLU\left( {{W_{C2}}{\kappa ^{nd}}} \right)} \right)
\end{equation}
\end{small}where $s^{nd}$ denotes the scaling factor. ${W_{A2}}$ and ${W_{C2}}$ are the parameters of convolution layers, whose channel dimensions are set to ${D \mathord{\left/ {\vphantom {D 4}} \right. \kern-\nulldelimiterspace} 4}$ and $D$, respectively. The output of the second-order channel correlations block can be represented as follows:
\begin{small}
\begin{equation}
X_d^{nd} = s_d^{nd}{x_d}
\end{equation}
\end{small}where $s_d^{nd}$ and ${x_d}$ denote the scaling factor and the feature map in the $d$-th channel. After that, we combine the first-order channel correlations block and second-order channel correlations block to obtain the multi-order channel correlations, which can be illustrated as follows:
\begin{small}
\begin{equation}
\hat X = {X^{st}} + {X^{nd}}
\end{equation}
\vspace{-2em}
\end{small}
\\\indent By integrating the multi-order spatial correlations module and multi-order channel correlations module via the proposed IMCG model, the multi-order spatial correlations and the multi-order channel correlations can be utilized to enhance the geometrical constraints and context information for generating more robust and accurate heatmaps.
\subsection{Explicit Probability-based Boundary-adaptive Regression method}
The facial boundary information \cite{Wu2018LookAB} has been shown to be beneficial for facial landmark detection as almost all the landmarks are defined lying on the facial boundaries. By utilizing the facial boundary heatmaps, Wu et al. \cite{Wu2018LookAB} can effectively utilize the boundary information to enhance the shape constraints between landmarks and improve the accuracy of FLD. However, when faces are suffering from extremely large poses or heavy occlusions, it is hard to generate accurate boundary heatmaps and add effective shape constraints to the predicted landmarks, resulting in performance degradation. Therefore, in this paper, we propose an Explicit Probability-based Boundary-adaptive Regression (EPBR) method to explore how to use the boundary information for robust facial landmark detection. Specifically, by integrating the boundary heatmap regression and landmark heatmap regression via the proposed searching mechanism, the EPBR method is able to generate more accurate and effective boundary-adaptive landmark heatmaps and then explicitly add both boundary and semantically consistent constraints to the final predicted landmarks to improve the accuracy of FLD.
\\\indent In the probabilistic view, training a CNN-based landmark detector can be formulated as a likelihood maximization problem:
\begin{small}
\begin{equation}
\mathop {\max }\limits_W \Phi \left( W \right) = p\left( {g\left| {{\rm M};W} \right.} \right)
\end{equation}
\end{small}where $g \in {R^{2L}}$ are the coordinates of ground-truth landmarks and $L$ is the number of landmarks, $M$ denotes the input image and $W$ represents the parameters of neural networks. For heatmap regression-based FLD methods, one pixel value on the heatmap works as the confidence of one particular landmark at that pixel. Hence, the whole heatmap represents the probability distribution over the image. In order to introduce the boundary information during learning, Eq. (18) can be reformulated as:
\begin{small}
\begin{equation}
\begin{array}{l}
\mathop {\max }\limits_W \Phi \left( W \right) = p\left( {g,H,{\rm B}\left| {{\rm M};W} \right.} \right)\\
{\kern 1pt} {\kern 1pt} {\kern 1pt} {\kern 1pt} {\kern 1pt} {\kern 1pt} {\kern 1pt} {\kern 1pt} {\kern 1pt}  {\kern 1pt}  {\kern 1pt} {\kern 1pt} {\kern 1pt} {\kern 1pt} {\kern 1pt} {\kern 1pt} {\kern 1pt} {\kern 1pt} {\kern 1pt} {\kern 1pt} {\kern 1pt} {\kern 1pt} {\kern 1pt} {\kern 1pt} {\kern 1pt} {\kern 1pt} {\kern 1pt} {\kern 1pt} {\kern 1pt} {\kern 1pt} {\kern 1pt}  = p\left( {g\left| {H,B} \right.} \right)p\left( {H,{\rm B}\left| {{\rm M};W} \right.} \right)\\
{\kern 1pt} {\kern 1pt} {\kern 1pt} {\kern 1pt} {\kern 1pt} {\kern 1pt} {\kern 1pt} {\kern 1pt} {\kern 1pt} {\kern 1pt} {\kern 1pt} {\kern 1pt} {\kern 1pt} {\kern 1pt} {\kern 1pt} {\kern 1pt} {\kern 1pt} {\kern 1pt} {\kern 1pt} {\kern 1pt} {\kern 1pt} {\kern 1pt} {\kern 1pt} {\kern 1pt} {\kern 1pt} {\kern 1pt} {\kern 1pt} {\kern 1pt} {\kern 1pt} {\kern 1pt}  {\kern 1pt}  = p\left( {g\left| {H,B} \right.} \right)p\left( {H\left| {{\rm M};W} \right.} \right)p\left( {{\rm B}\left| {{\rm M};W} \right.} \right)
\end{array}
\end{equation}
\end{small}where $B$ denotes the boundary heatmap and $H$ represents the landmark heatmap. We aim to generate more accurate and effective boundary-adaptive landmark heatmaps by designing and optimizing $\Phi \left( W \right)$.
\subsubsection{Landmark Heatmap}
The ground-truth landmark heatmap is generated by a two-dimensional Gaussian distribution which can be illustrated as:
\begin{small}
\begin{equation}
{H^*}\left( {x,y} \right) = \exp \left( { - \frac{{{{\left( {x - u} \right)}^2} + {{\left( {y - v} \right)}^2}}}{{2\sigma _1^2}}} \right)
\end{equation}
\end{small}where $\left( {u,v} \right) \in {S^*}$ and ${S^*}$ denotes the ground-truth face shape, $\left( {u,v} \right)$ represents the coordinates of a landmark in ${S^*}$. ${H^*}$ denotes the ground-truth landmark heatmap, while $\left( {x,y} \right)$ means a pixel in ${H^*}$. For any position $\left( {x,y} \right)$, the more the heatmap region around $\left( {x,y} \right)$ follows a standard Gaussian, the more the position $\left( {x,y} \right)$ is likely to be $\left( {u,v} \right)$. Therefore, we can use the distribution distance between the predicted and the ground-truth landmark heatmap to model $p\left( {H\left| {{\rm M};W} \right.} \right)$, that is, the Jensen-Shannon divergence loss (JSDL) is used. Compared to the mean squared error, 1) JSDL can pay more attention to the foreground area of heatmaps rather than treating the whole heatmap equally, 2) JSDL can accurately measure the difference between two distributions. Therefore, JSDL can be utilized to generate more effective heatmaps and better fit deep neural networks, which can be stated as:
\begin{small}
\begin{equation}
\begin{array}{l}
JS\left( {{p_{{H^*}}}||{p_{H}}} \right) = \frac{1}{2}KL\left( {{p_{H}}\left( {x,y} \right)||\frac{{{p_{H}}\left( {x,y} \right) + {p_{{H^*}}}\left( {x,y} \right)}}{2}} \right)\\
 + \frac{1}{2}KL\left( {{p_{{H^*}}}\left( {x,y} \right)||\frac{{{p_{H}}\left( {x,y} \right) + {p_{{H^*}}}\left( {x,y} \right)}}{2}} \right)
\end{array}
\end{equation}
\end{small}where ${p_{H}}$ and ${p_{{H^*}}}$ denote the distributions of the generated and the ground-truth landmark heatmaps, respectively. $KL$ means the Kullback-Leibler Divergence loss. After that, the joint probability $p\left( {H\left| {{\rm M};W} \right.} \right)$ can be modeled as follows:
\begin{small}
\begin{equation}
p\left( {H\left| {{\rm M};W} \right.} \right) = \exp \left( { - \sum\nolimits_\ell  {JS\left( {p_{{H^*}}^\ell ||p_{H}^\ell } \right)} } \right)
\end{equation}
\end{small}where $\ell $ denotes the landmark index. By optimizing Eq. (22), we can generate more accurate and effective landmark heatmaps.
\begin{figure}[t]
\begin{center}
\includegraphics[width=0.9\linewidth]{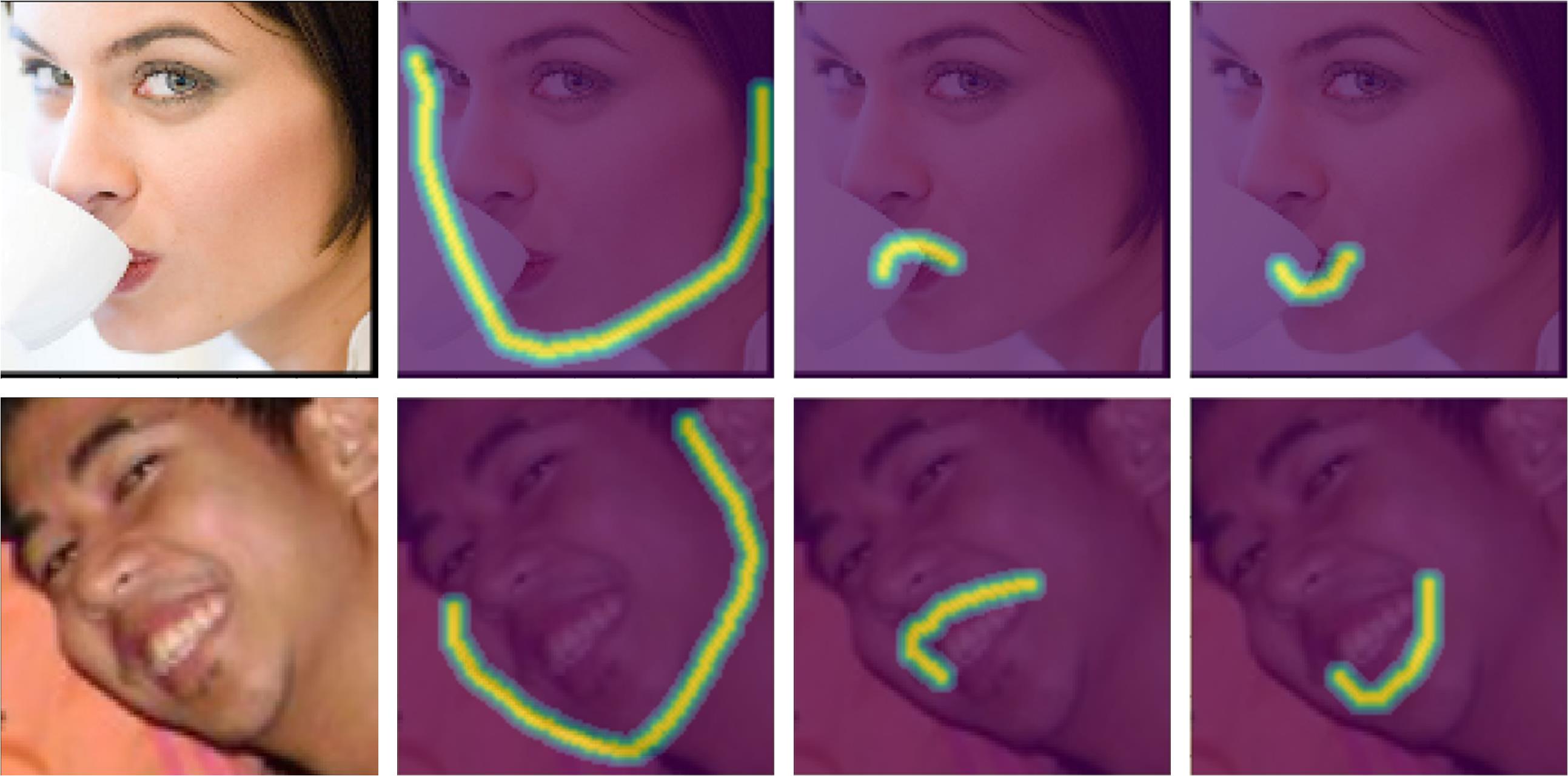}
\end{center}
\vspace{-1em}
   \caption{Boundary heatmap. The second, third, and fourth columns represent the corresponding facial boundary heatmaps of the facial outer contour, the upper side of the upper lip and the lower side of the lower lip, respectively. With such well-defined facial boundaries, we can effectivly model the geometric structure of the human face to enhance the shape constraints between landmarks for robust FLD under large poses and heavy occlusions.}
   \vspace{-1em}
\label{fig5}
\end{figure}
\subsubsection{Boundary Heatmap}
Following LAB \cite{Wu2018LookAB}, there are 13 boundary heatmaps which contain the corresponding landmarks. For each boundary, landmarks on this boundary are firstly interpolated to get a dense boundary line and then transformed into a ground-truth boundary heatmap by using a Gaussian distribution with a standard deviation ${\sigma _2}$. Specifically, we firstly use a binary map and a distance transform function to get distance map $Dist$, and then transform the distance map $Dist$ into a ground-truth boundary heatmap (see Fig. \ref{fig5}). With such facial boundary heatmaps, the geometric structure of the human face can be identified easily, and more accurate landmarks will be detected. The whole process of generating the ground-truth boundary heatmap with the distance map $Dist$ can be illustrated as follows:
\begin{small}
\begin{equation}
B_t^*\left( {x,y} \right) = \left\{ {\begin{array}{*{20}{c}}
	{\exp \left( { - \frac{{Dis{t_t}\left( {x,y} \right)}}{{2\sigma _2^2}}} \right){\kern 1pt} {\kern 1pt} {\kern 1pt} {\kern 1pt} {\kern 1pt} {\kern 1pt} ,if{\kern 1pt} Dis{t_t}\left( {x,y} \right) < 2{\sigma _2}{\kern 1pt} }\\
	{\xi ,{\kern 1pt} {\kern 1pt} {\kern 1pt} {\kern 1pt} {\kern 1pt} {\kern 1pt} {\kern 1pt} {\kern 1pt} {\kern 1pt} {\kern 1pt} {\kern 1pt} {\kern 1pt} {\kern 1pt} {\kern 1pt} {\kern 1pt} {\kern 1pt} {\kern 1pt} {\kern 1pt} {\kern 1pt} {\kern 1pt} {\kern 1pt} {\kern 1pt} {\kern 1pt} {\kern 1pt} {\kern 1pt} {\kern 1pt} {\kern 1pt} {\kern 1pt} {\kern 1pt} {\kern 1pt} {\kern 1pt} {\kern 1pt} {\kern 1pt} {\kern 1pt} {\kern 1pt} {\kern 1pt} {\kern 1pt} {\kern 1pt} {\kern 1pt} {\kern 1pt} {\kern 1pt} {\kern 1pt} {\kern 1pt} {\kern 1pt} {\kern 1pt} {\kern 1pt} {\kern 1pt} {\kern 1pt} {\kern 1pt} {\kern 1pt} {\kern 1pt} {\kern 1pt} {\kern 1pt} {\kern 1pt} {\kern 1pt} {\kern 1pt} {\kern 1pt} {\kern 1pt} {\kern 1pt} {\kern 1pt} {\kern 1pt} {\kern 1pt} {\kern 1pt} {\kern 1pt} {\kern 1pt} {\kern 1pt} {\kern 1pt} {\kern 1pt} {\kern 1pt} {\kern 1pt} {\kern 1pt} {\kern 1pt} {\kern 1pt} {\kern 1pt} {\kern 1pt} {\kern 1pt} {\kern 1pt} {\kern 1pt} {\kern 1pt} {\kern 1pt} {\kern 1pt} {\kern 1pt} {\kern 1pt} {\kern 1pt} {\kern 1pt} {\kern 1pt} {\kern 1pt} {\kern 1pt} {\kern 1pt} {\kern 1pt} {\kern 1pt} {\kern 1pt} {\kern 1pt} {\kern 1pt} {\kern 1pt} {\kern 1pt} {\kern 1pt} {\kern 1pt} {\kern 1pt} {\kern 1pt} {\kern 1pt} otherwise}
	\end{array}} \right.
\end{equation}
\end{small}
\\where $\xi $ is a small constant, $t = 1 \cdots T$ means the boundary index and $T=13$. We also use the JSDL to measure the distribution difference which can be stated as:
\begin{small}
\begin{equation}
\begin{array}{l}
JS\left( {{p_{{B^*}}}||{p_{B}}} \right) = \frac{1}{2}KL\left( {{p_{B}}\left( {x,y} \right)||\frac{{{p_{B}}\left( {x,y} \right) + {p_{{B^*}}}\left( {x,y} \right)}}{2}} \right)\\
 + \frac{1}{2}KL\left( {{p_{{B^*}}}\left( {x,y} \right)||\frac{{{p_{B}}\left( {x,y} \right) + {p_{{B^*}}}\left( {x,y} \right)}}{2}} \right)
\end{array}
\end{equation}
\end{small}where ${p_{B}}$ and ${p_{{B^*}}}$ denote the distributions of the generated and the ground-truth boundary heatmaps, respectively. After that, the joint probability can also be modeled as a product of boundary heatmap similarity maximized over all boundaries:
\begin{small}
\begin{equation}
p\left( {B\left| {{\rm M};W} \right.} \right) = \exp \left( { - \sum\nolimits_t {JS\left( {p_{{B^*}}^t||p_{B}^t} \right)} } \right)
\end{equation}
\vspace{-1em}
\end{small}
\subsubsection{Boundary-adaptive Landmark Heatmap}
After generating landmark heatmaps and boundary heatmaps, we can predict landmark $\tilde g$ by traversing the generated landmark heatmaps and the accuracy of predicted landmark $\tilde g$ is improved by using the boundary information. However, when faces are suffering from large poses and partial occlusions, the predicted landmark $\tilde g$ may deviate from the corresponding facial boundary (i.e., not on the corresponding facial boundary) or landmark (on the corresponding facial boundary, but far away from the ground-truth landmark $g$). As the boundary heatmaps are more robust to partial occlusions and large poses for their continuous characteristics, we firstly construct boundary-adaptive landmark heatmaps $\hat {H}$ and then search the optimal landmark $\hat g$ around the predicted landmark $\tilde g$ in $\hat {H}$. $\hat {H}$ can be formulated as follows:
\begin{small}
\begin{equation}
{\hat {H}^\ell }\left( {x,y} \right) = {{H}^\ell }\left( {x,y} \right) \otimes {{B}^t}\left( {x,y} \right)
\end{equation}
\end{small}where ${\hat {H}^\ell }$ means the boundary-adaptive landmark heatmap of the $\ell$-th landmark, ${B^t}$ represents the corresponding boundary heatmap and $\otimes $ means element-wise multiplication. By the $\otimes $ operation, the boundary information is ready to improve the accuracy of the predict landmarks.
\subsubsection{Searching Mechanism}
For the ground-truth boundary-adaptive landmark heatmaps, $\hat g$, $\tilde g$ and $g$ should represent the same landmark. However, when faced with large poses and heavy occlusions, the predicted landmark $\tilde g$ usually deviates from the ground-truth landmark $g$. Hence, a searching mechanism is proposed to firstly search for the optimal landmark $\hat g$ around $\tilde g$ in ${\hat {H}}$, and then the error between $\hat g$ and $\tilde g$ is used to fit the neural network so that semantically consistent boundary-adaptive landmark heatmaps can be generated. Therefore, the proposed searching mechanism should make the optimal $\hat g$ meet the following two criteria. First, $\hat g$ should have larger confidence in the generated boundary-adaptive landmark heatmap, which corresponds to $p\left( {H,{\rm B}\left| {{\rm M};W} \right.} \right)$. Second, $\hat g$ should be close to $\tilde g$ (i.e., $\hat g$, $\tilde g$ and $g$ should represent the same landmark), which corresponds to $p\left( {g\left| {H,B} \right.} \right)$. Hence, $p\left( {g\left| {H,B} \right.} \right)$ can be defined as the Gaussian similarity over all $\left\{ {{{\hat g}^\ell },{{\tilde g}^\ell }} \right\}$ pairs:
\begin{small}
\begin{equation}
\begin{array}{l}
p\left( {g\left| {H,B} \right.} \right) \propto \prod\limits_\ell  {\exp \left( { - \frac{{{{\left\| {{{\tilde g}^\ell } - {{\hat g}^\ell }} \right\|}^2}}}{{2\sigma _3^2}}} \right)} \\
{\kern 1pt} {\kern 1pt} {\kern 1pt} {\kern 1pt} {\kern 1pt} {\kern 1pt} {\kern 1pt} {\kern 1pt} {\kern 1pt} {\kern 1pt} {\kern 1pt} {\kern 1pt} {\kern 1pt} {\kern 1pt} {\kern 1pt} {\kern 1pt} {\kern 1pt} {\kern 1pt} {\kern 1pt} {\kern 1pt} {\kern 1pt} {\kern 1pt} {\kern 1pt} {\kern 1pt} {\kern 1pt} {\kern 1pt} {\kern 1pt} {\kern 1pt} {\kern 1pt} {\kern 1pt} {\kern 1pt} {\kern 1pt} {\kern 1pt} {\kern 1pt} {\kern 1pt} {\kern 1pt} {\kern 1pt} {\kern 1pt} {\kern 1pt} {\kern 1pt} {\kern 1pt} {\kern 1pt} {\kern 1pt} {\kern 1pt} {\kern 1pt} = \exp \left( {\sum\limits_\ell  { - \frac{{{{\left\| {{{\tilde g}^\ell } - {{\hat g}^\ell }} \right\|}^2}}}{{2\sigma _3^2}}} } \right)
\end{array}
\end{equation}
\vspace{-0.5em}
\end{small} 
\\\indent With the proposed searching mechanism, we add a new and clear constraint to the network so that it can adaptively adjust parameters to generate more accurate and consistent landmark and boundary heatmaps. Then, by incorporating the JSDL with the searching mechanism, both boundary and semantically consistent constraints can be explicitly added to the predicted landmarks to improve the accuracy of FLD. Therefore, we name it \textbf{Explicit Probability-based Boundary-adaptive Regression (EPBR) method}.
\\\indent Finally, by integrating $p\left( {o\left| {H,B} \right.} \right)$, $p\left( {H\left| {{\rm M};W} \right.} \right)$ and $p\left( {B\left| {{\rm M};W} \right.} \right)$ in one framework and take log function, the objective function of the EPBR method is formulated as follows:
\begin{small}
	\begin{equation}
\begin{array}{l}
\mathop {{\rm{min}}}\limits_{\hat g,W}  - \log \Phi \left( {\hat g,W} \right) = \sum\limits_{t = 1}^T {JS\left( {p_{{B^*}}^t\left\| {p_B^t} \right.} \right)} \\
{\rm{ + }}\sum\limits_{\ell  = 1}^L {\left( {\eta \frac{{{{\left\| {{{\tilde g}^\ell } - {{\hat g}^\ell }} \right\|}^2}}}{{2\sigma _3^2}}{\rm{ + }}JS\left( {p_{{H^*}}^\ell \left\| {p_H^\ell } \right.} \right)} \right)}
\end{array}
	\end{equation}
\end{small}where $\eta$ denotes the weight corresponding to $p\left( {g\left| {H,B} \right.} \right)$. By incorporating the JSDL with the searching mechanism, the EPBR method can be used to better fit the neural networks and generate more effective boundary-adaptive landmark heatmaps for more robust FLD.
\begin{table*}
	\centering
	\caption{The detailed network structure of the proposed MMDN}
	\begin{tabular}{p{2.4cm}p{2cm}p{2cm}p{6cm}}
		\hline
		Layer & Shape\_in & Shape\_out & Kernel/Stride \\
		\hline
		input & - & $256\times256\times3$ & -  \\
		conv/batch\_norm & $256\times256\times3$ & $128\times128\times64$ & $7\times7\times64/2$ \\
		residual block &$128\times128\times64$ & $128\times128\times128$ & $1\times1\times32/1$, $3\times3\times32/1$, $1\times1\times128/1$  \\
		avg\_pooling & $128\times128\times128$ & $64\times64\times128$ & $2\times2/2$ \\
		residual block$\times  2$ & $64\times64\times128$ & $64\times64\times128$ & $1\times1\times32/1$, $3\times3\times32/1$, $1\times1\times128/1$ \\
		conv   &$64\times64\times128$ & $64\times64\times256$ & $3\times3\times256/1$ \\\hline
		\multicolumn{4}{l}{\textbf{multi-order spatial correlations module}}\\
		branch1  & $64\times64\times256$ & $64\times64\times256$ & $1\times1\times64/1$, $3\times3\times64/1$, $1\times1\times256/1$ \\
		max\_pooling1 & $64\times64\times256$ & $32\times32\times256$ & $2\times2/2$ \\
		residual block & $32\times32\times256$ & $32\times32\times256$ & $1\times1\times64/1$, $3\times3\times64/1$, $1\times1\times256/1$ \\
		branch2  & $32\times32\times256$ & $32\times32\times256$ & $1\times1\times64/1$, $3\times3\times64/1$, $1\times1\times256/1$ \\
		max\_pooling2 & $32\times32\times256$ & $16\times16\times256$ & $2\times2/2$ \\
		residual block & $16\times16\times256$ & $16\times16\times256$ & $1\times1\times64/1$, $3\times3\times64/1$, $1\times1\times256/1$ \\
		branch3  & $16\times16\times256$ & $16\times16\times256$ & $1\times1\times64/1$, $3\times3\times64/1$, $1\times1\times256/1$ \\
		max\_pooling3 & $16\times16\times256$ & $8\times8\times256$ & $2\times2/2$ \\
		residual block & $8\times8\times256$ & $8\times8\times256$ & $1\times1\times64/1$, $3\times3\times64/1$, $1\times1\times256/1$ \\
		branch4  & $8\times8\times256$ & $8\times8\times256$ & $1\times1\times64/1$, $3\times3\times64/1$, $1\times1\times256/1$ \\
		max\_pooling4 & $8\times8\times256$ & $4\times4\times256$ & $2\times2/2$ \\
		residual block$\times  3$ & $4\times4\times256$ & $4\times4\times256$ & $1\times1\times64/1$, $3\times3\times64/1$, $1\times1\times256/1$ \\
		upsampling4 & $4\times4\times256$ & $8\times8\times256$ & - \\
		add\_branch4 &$8\times8\times256$ & $8\times8\times256$ & - \\
		residual block & $8\times8\times256$ & $8\times8\times256$ & $1\times1\times64/1$, $3\times3\times64/1$, $1\times1\times256/1$ \\
		upsampling3 & $8\times8\times256$ & $16\times16\times256$ & - \\
		add\_branch3 & $16\times16\times256$ & $16\times16\times256$ & - \\
		residual block & $16\times16\times256$ & $16\times16\times256$ & $1\times1\times64/1$, $3\times3\times64/1$, $1\times1\times256/1$ \\
		upsampling2 & $16\times16\times256$ & $32\times32\times256$ & - \\
		add\_branch2 &$32\times32\times256$ & $32\times32\times256$ & - \\
		residual block & $32\times32\times256$ & $32\times32\times256$ & $1\times1\times64/1$, $3\times3\times64/1$, $1\times1\times256/1$ \\
		upsampling1 & $32\times32\times256$ & $64\times64\times256$ & - \\
		add\_branch1 &$64\times64\times256$ & $64\times64\times256$ & - \\
		residual block (Q) & $64\times64\times256$ & $64\times64\times256$ & $1\times1\times64/1$, $3\times3\times64/1$, $1\times1\times256/1$ \\
		$P' = \lambda \hat OQ + Q $ & - & $64\times64\times256$ & - \\\hline
		\multicolumn{4}{l}{\textbf{multi-order channel correlations module}}\\
		$X$ & - & $64\times64\times256$ & - \\
		GAP & $64\times64\times256$ & $1\times1\times256$ & - \\
		conv1\_1 & $1\times1\times256$ & $1\times1\times64$ & $1\times1\times64$/1 \\
		conv1\_2 ($s^{st}$) & $1\times1\times64$ & $1\times1\times256$ & $1\times1\times256$/1 \\
		GCP & $64\times64\times256$ & $1\times1\times256$ & - \\
		conv2\_1 & $1\times1\times256$ & $1\times1\times64$ & $1\times1\times64$/1 \\
		conv2\_2 ($s^{nd}$) & $1\times1\times64$ & $1\times1\times256$ & $1\times1\times256$/1 \\
		\multicolumn{4}{l}{\textbf{$X = X{s^{st}} + X{s^{nd}}$}}\\\hline
		deconv & $64\times64\times256$ & $128\times128\times128$ & $4\times4\times128$/2 \\
		out & $128\times128\times128$ & $128\times128\times81$ & $3\times3\times81$/1 \\
		\hline
	\end{tabular}
\vspace{-1em}
	\label{table1}
\end{table*}
\subsection{Multi-order Multi-constraint Deep Networks}
The proposed Multi-order Correlating Geometry-aware (IMCG) model can obtain more discriminative representations by introducing the high-order information (i.e., the multi-order spatial correlations and the multi-order channel correlations) to implicitly enhance the geometric constraints and context information. Then, the Explicit Probability-based Boundary-adaptive Regression (EPBR) method is proposed to incorporate the boundary heatmap regression with landmark heatmap regression and further explicitly add both boundary constraints and semantically consistent constraints to the final predicted landmarks. Finally, by integrating the IMCG model and the EPBR method into a Multi-order Multi-constraint Deep Network (MMDN) via a seamless formulation, we can generate more accurate landmark heatmaps and boundary heatmaps that can be further used to improve the accuracy of FLD under large poses and heavy occlusions. The detailed network structure of the proposed MMDN is shown in Table \ref{table1}.
\subsection{Optimization}
Combining Eq. (18), (19), (22), (25), (27) and (28), the objective function of MMDN can be reformulated as:
\begin{small}
\begin{equation}
\begin{array}{l}
\mathop {{\rm{min}}}\limits_{\hat g,W}  - \log \Phi \left( {\hat g,W} \right) = \sum\limits_{t = 1}^T {JS\left( {p_{{B^*}}^t\left\| {p_{B}^t} \right.} \right)} \\ {\kern 1pt} {\kern 1pt} {\kern 1pt} {\kern 1pt} {\kern 1pt} {\kern 1pt} {\kern 1pt} {\kern 1pt} {\kern 1pt} {\kern 1pt} {\kern 1pt} {\kern 1pt} {\kern 1pt} {\kern 1pt} {\kern 1pt}
{\rm{ + }}\sum\limits_{\ell  = 1}^L {\left( {\eta \frac{{{{\left\| {{{\tilde g}^\ell } - {{\hat g}^\ell }} \right\|}^2}}}{{2\sigma _3^2}}{\rm{ + }}JS\left( {p_{{H^*}}^\ell \left\| {p_{H}^\ell } \right.} \right)} \right)} \\
= \sum\limits_{t = 1}^T {JS\left( {p_{{B^*}}^t\left\| {MMDN{{\left( {M,W} \right)}^t}} \right.} \right)} \\
+ \sum\limits_{\ell  = 1}^L {\left( {\eta \frac{{{{\left\| {{{\tilde g}^\ell } - {{\hat g}^\ell }} \right\|}^2}}}{{2\sigma _3^2}}{\rm{ + }}JS\left( {p_{{H^*}}^\ell \left\| {MMDN{{\left( {M,W} \right)}^\ell }} \right.} \right)} \right)}
\end{array}
\end{equation}
\end{small}where $W$ denotes the parameters of the proposed MMDN. By inputting a face image, MMDN can output the corresponding landmark heatmaps and boundary heatmaps. Then, we construct and search the boundary-adaptive landmark heatmaps to obtain the optimal landmarks.
\\\indent Since the proposed MMDN contains multiple variables, we design an iterative algorithm to alternately optimize them. In each iteration, firstly, the parameter $W$ of MMDN is fixed, the optimal landmarks can be predicted according to the boundary-adaptive landmark heatmaps. Then the predicted optimal landmarks $\hat g$ are fixed and $W$ is updated under the supervision of $- \log \Phi \left( {\hat g,W} \right)$. The whole optimization process can be iterated alternately in the following two steps.
\\\indent Step 1: Given $W$, we have the following sub-problem:
\begin{small}
\begin{equation}
\mathop {\max }\limits_{{\hat g}^l} \left( {\exp \left( {\frac{{{{\left\| {{{\tilde g}^\ell } - {{\hat g}^\ell }} \right\|}^2}}}{{ - 2\sigma _3^2}}} \right){\rm{ + }}p_{H}^\ell \left( {{{\hat g}^l}} \right)p_{B}^t\left( {{{\hat g}^l}} \right)} \right)
\end{equation}
\end{small}
\\where all the variables are known except the ${\hat g^\ell }$. We search ${\hat g^\ell }$ by going through all the pixels around ${\tilde g^\ell }$, and the one with the maximum value in Eq. (30) is the solution.
\\\indent  Step 2: When the optimal landmarks are fixed, we have the following sub-problem:
\begin{small}
\begin{equation}
\begin{array}{l}
\mathop {{\rm{min}}}\limits_W \sum\limits_{t = 1}^T {JS\left( {p_{{B^*}}^t\left\| {MMDN{{\left( {M,W} \right)}^t}} \right.} \right)} \\
+ \sum\limits_{\ell  = 1}^L {\left( {\eta \frac{{{{\left\| {{{\tilde g}^\ell } - {{\hat g}^\ell }} \right\|}^2}}}{{2\sigma _3^2}}{\rm{ + }}JS\left( {p_{{H^*}}^\ell \left\| {MMDN{{\left( {M,W} \right)}^\ell }} \right.} \right)} \right)}
\end{array}
\end{equation}
\end{small}
\\\indent The optimization is a typical network training process under the supervision of Eq. (31). With the proposed IMCG model, we can obtain more representative and discriminative features. Then, by integrating the IMCG model and the EPBR method into the MMDN, the optimal landmarks can be easily predicted, which helps achieve more robust FLD.
\section{Experiments}
In this section, we firstly introduce the evaluation settings including the datasets and methods for comparison. Then, we compare our algorithm with the state-of-the-art FLD methods on challenging benchmark datasets such as 300W \cite{Sagonas2016300FI}, COFW \cite{Burgosartizzu2013Robust}, AFLW \cite{Zhu2016UnconstrainedFA} and WFLW \cite{Wu2018LookAB}.
\subsection{Dataset and Implementation details}
\indent\textbf{300W }(68 landmarks): It consists of some present datasets including LFPW \cite{Zhu2012FaceDP}, AFW \cite{Belhumeur2011LocalizingPO}, Helen \cite{le2012interactive}, and IBUG \cite{Sagonas2016300FI}. With a total of 3148 pictures, the training sets are made up of the training sets of AFW, LFPW and Helen while the testing set includes 689 images with such testing sets as IBUG, LFPW and Helen. The testing set is further divided  into three subsets: 1) Challenging subset (i.e., IBUG dataset). It contains 135 images that are collected from a more general ``in the wild" scenarios, and experiments on IBUG dataset are more challenging. 2) Common subset (554 images, including 224 images from LFPW test set and 330 images from Helen test set). 3) Fullset (689 images, containing the challenging subset and common subset).
\\\indent\textbf{COFW }(29 landmarks): It is another very challenging dataset on occlusion issues which is published by Burgos-Artizzu et al. \cite{Burgosartizzu2013Robust}. It contains 1345 training images and the testing set of COFW contains 507 face images under heavy occlusions, large pose and expression variations.
\\\indent\textbf{AFLW }(19 landmarks): It contains 25993 face images with jaw angles ranging from ${\rm{ - }}{120^ \circ }$ to ${\rm{ + }}{120^ \circ }$ and pitch angles ranging from ${\rm{ - }}{90^ \circ }$ to ${\rm{ + }}{90^ \circ }$. Meantime, these pictures are very different in pose and occlusion. AFLW-full divides the 24386 images into two parts: 20000 for training and 4386 for testing. In the meantime, AFLW-frontal selects 1165 images out of the 4386 testing images to evaluate the alignment algorithm on frontal faces.
\\\indent\textbf{WFLW} (98 landmarks): It contains 10000 faces (7500 for training and 2500 for testing) with 98 landmarks. Apart from landmark annotation, WFLW also has rich attribute annotations (such as occlusion, pose, make-up, illumination, blur and expression.) that can be used for a comprehensive analysis of existing algorithms.
\\\indent\textbf{Evaluation metric} Normalized Mean Error (NME) \cite{Belhumeur2011LocalizingPO,cao2014face} is commonly used to evaluate FLD methods. First, the error between the predicted positions and the ground-truth is calculated, then it is further normalized to avoid larger errors in higher resolution images. For the 300W \cite{Sagonas2016300FI} and the COFW \cite{Burgosartizzu2013Robust}, NME normalized by the inter-pupil distance is used. For the AFLW \cite{Zhu2016UnconstrainedFA}, we use NME normalized by the face size. For the WFLW  \cite{Wu2018LookAB}, NME normalized by inter-ocular distance is adopted.
\\\indent\textbf{Implementation Details} In our experiments, all the training and testing images are cropped and resized to $256 \times 256$ according to the provided bounding boxes. To perform data augmentation, we randomly sample the angle of rotation ($-60^\circ, +60^\circ$) and the bounding box scale ($0.8, 1.2$) from a standard Gaussian distribution. We use the Hourglass Network \cite{Yang2017StackedHN} as our backbone to construct the proposed Multi-order Multi-constraint Deep Network, and the output heatmap size is $128 \times 128$. The training of MMDN takes 100000 iterations and the staircase function is used to set the learning rate. The initial learning rate is $2.5 \times {10^{{\rm{ - 4}}}}$ and then it is divided by 5, 2 and 2 at iteration 5000, 20000 and 50000, respectively. ${\sigma _1}$ and ${\sigma _2}$ are set to 3, ${\sigma _3}$ and $\eta$ are set to 4 and 10 respectively. During the searching of the optimal landmark $\hat g$, we predict $\hat g$ from a $7 \times 7$ region centered on the landmark $\tilde g$ in the generated boundary-adaptive landmark heatmap. The MMDN is trained with Pytorch on 8 Nvidia Tesla V100 GPUs.
\\\indent\textbf{Experiment Settings} To better evaluate the effectiveness of each module proposed in this paper, we firstly use the Stacked Hourglass Network (SHN) \cite{Yang2017StackedHN}  as the baseline, and then we separately combine SHN with the proposed multi-order spatial correlations (MSC) module, multi-order channel correlations (MCC) module, the IMCG model and the EPBR method. We also combine the SHN with the first-order channel correlations (FCC) block to test the effectiveness of channel re-calibration. The detailed experimental results are shown below.
\subsection{Evaluation under Normal Circumstances}
Faces in the 300W common subset and 300W full set have fewer variations on the head pose, facial expression and occlusion. Therefore, we evaluate the effectiveness of our method under normal circumstances on these two subsets. Table \ref{tab1} shows the experimental results in comparison to the existing benchmarks. From Table \ref{tab1}, we can see that our method achieves 3.17\% NME on the 300W common subset and 3.74\% NME on the 300W full set, which outperforms state-of-the-art methods on faces under normal circumstances. These results indicate that our MMDN can improve the accuracy of FLD under normal circumstances, mainly because 1) the IMCG model can obtain more discriminative representations by introducing the multi-order spatial and channel correlations to enhance the geometrical constraints. 2) the EPBR method can effectively incorporate the boundary heatmap regression with landmark heatmap regression and further explicitly add both boundary and semantically consistent constraints to the predicted landmarks.
\begin{table}
\caption{Comparisons with state-of-the-art methods on the 300W dataset. The error (NME) is normalized by the inter-pupil distance. (\% omitted)}
\begin{center}
\begin{tabular}{p{3.8cm}p{0.8cm}p{1.2cm}p{0.8cm}}
\hline
Method & Commom  & Challenging  & Full  \\
\hline
LBF \cite{ren2014face} &	4.95 &	11.98 &	6.32\\
RAR(ECCV16) \cite{Xiao2016Robust}&	4.12	&8.35	&4.94\\
DCFE(ECCV18) \cite{Valle2018ADC}&	3.83&	7.54&	4.55\\
Wing(CVPR18) \cite{Feng2017WingLF}&	3.27&	7.18&	4.04\\
LAB(CVPR18) \cite{Wu2018LookAB}&	3.42	&6.98&	4.12\\
SBR(CVPR18) \cite{Dong2018SupervisionbyRegistrationAU}&	3.28&	7.58&	4.10\\
Liu et al. (CVPR19) \cite{Liu2019SemanticAF} &	3.45&	6.38&	4.02\\
ODN(CVPR19) \cite{Zhu2019RobustFL}	&3.56	&6.67&	4.17\\
\hline
SHN &	4.43&	7.56&	5.04\\
SHN+FCC &	4.22&	7.26&	4.82\\
SHN+MCC &	4.07&	7.04&	4.65\\
SHN+MSC &   3.67&	6.63&	4.25\\
SHN+IMCG&	3.43&	6.38&	4.01\\
SHN+EPBR&	3.82&	6.97&	4.44\\
\textbf{SHN+IMCG+EPBR (MMDN)}&	\textbf{3.17}&	\textbf{6.08}&	\textbf{3.74}\\
\hline
\end{tabular}
\end{center}
\vspace{-2em}
\label{tab1}
\end{table}
\begin{figure}[t]
\begin{center}
\includegraphics[width=0.92\linewidth]{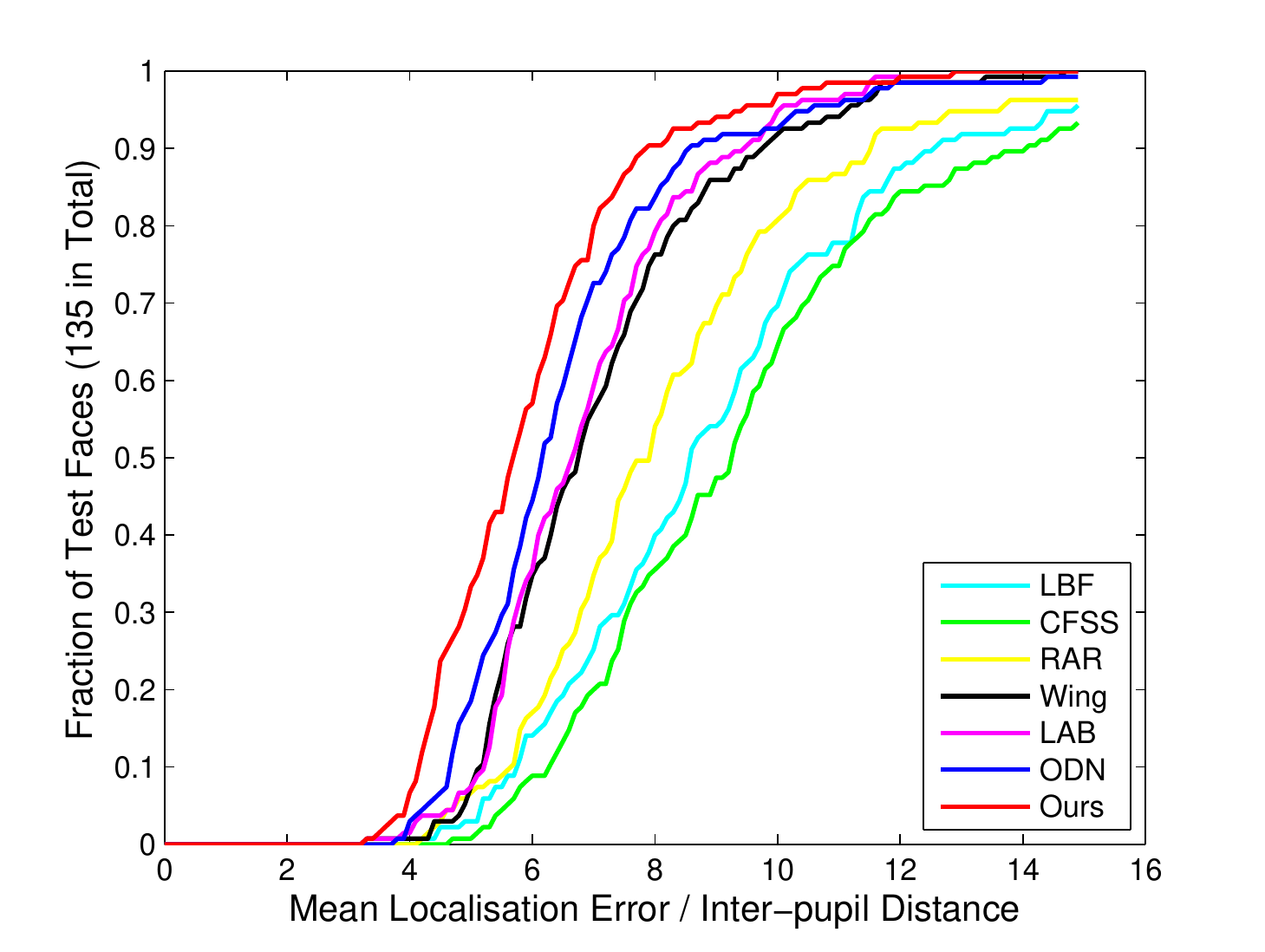}
\end{center}
   \caption{Comparison of CED curves between our method and state-of-the-art methods including LBF \cite{ren2014face}, CFSS \cite{zhu2015face}, RAR \cite{Xiao2016Robust}, Wing \cite{Feng2017WingLF}, LAB \cite{Wu2018LookAB} and ODN \cite{Zhu2019RobustFL} on the 300W Challenging subset (68 landmarks). Our approach is more robust to partial occlusions and large poses than other methods. Best viewed in color.}
   \vspace{-1em}
\label{fig6}
\end{figure}
\begin{table}
\caption{Comparisons with state-of-the-art methods on the COFW dataset. The error (NME) is normalized by the inter-pupil distance. (- not counted, \% omitted)}
\vspace{-1em}
\begin{center}
\begin{tabular}{p{3.8cm}p{1.6cm}p{1.6cm}}
\hline
Method & NME  & Failure \\
\hline
human&	5.6&	-\\
PCPR \cite{Burgosartizzu2013Robust}&	8.50&	20.00\\
HPM \cite{Ghiasi2014Occlusion}&	7.50&	13.00\\
CCR \cite{Feng2015Cascaded}&	7.03&	10.9\\
DRDA \cite{Zhang2016OcclusionFreeFA}&	6.46&	6.00\\
RAR \cite{Xiao2016Robust}&	6.03&	4.14\\
DAC-CSR(CVPR17) \cite{Feng2017DynamicAC}&	6.03&	4.73\\
CAM \cite{Wan2019FaceAB}&	5.95&	3.94\\
CRD \cite{wan2020robust}&	5.72&	3.76\\
LAB(CVPR18) \cite{Wu2018LookAB}&	5.58&	2.76\\
ODN(CVPR19) \cite{Zhu2019RobustFL}&	5.30&	-\\
\hline
SHN &	6.21&	5.52\\
SHN+FCC &	6.01&	4.54\\
SHN+MCC &	5.86&	3.75\\
SHN+MSC &   5.52&	2.76\\
SHN+IMCG&	5.32&	2.17\\
SHN+EPBR&	5.62&	3.16\\
\textbf{SHN+IMCG+EPBR (MMDN)} &	\textbf{5.01}&	\textbf{1.78} \\
\hline
\end{tabular}
\end{center}
\vspace{-2em}
\label{tab2}
\end{table}
\begin{table*}
	\caption{Comparisons with state-of-the-art methods on WFLW dataset.NME is normalized by the inter-ocular distance. (\% omitted)}
	\vspace{-1em}
	\begin{center}
		\begin{tabular}{p{3.8cm}p{1.5cm}p{1.5cm}p{1.5cm}p{1.5cm}p{1.5cm}p{1.5cm}p{1.5cm}}
			\hline
			Method & Testset  & Pose Subset  & Expression Subset &Illumination Subset &Make-Up Subset &Occlusion Subset & Blur Subset  \\
			\hline
			ESR \cite{cao2014face} &11.13 &25.88 &11.47 &10.49 &11.05 &13.75 &12.20 \\
			SDM \cite{xiong2013supervised} &10.29 &24.10 &11.45 &9.32 &9.38 &13.03 &11.28 \\
			CCFS(CVPR15) \cite{zhu2015face} &9.07 &21.36 &10.09 &8.30 &8.74 &11.76 &9.96 \\
			LAB(CVPR18) \cite{Wu2018LookAB} &5.27 &10.24 &5.51 &5.23 &5.15 &6.79 &6.32 \\
			Wing(CVPR18) \cite{Feng2017WingLF} &5.11 &8.75 &5.36 &4.93 &5.41 &6.37 &5.81 \\			
			\hline
			SHN &6.78 &14.42 &7.19 &6.13 &6.21 &7.97 &7.64 \\
			SHN+IMCG &5.25 &8.32 &5.21 &4.87 &4.98 &6.42 &6.01 \\
			SHN+EPBR &5.46 &8.55 &5.46 &5.15 &5.21 &6.76 &6.39 \\	
			\textbf{SHN+IMCG+EPBR (MMDN)} &\textbf{4.87} &\textbf{8.15} &\textbf{4.99} &\textbf{4.61} &\textbf{4.72} &\textbf{6.17} &\textbf{5.72}\\
			\hline
		\end{tabular}
	\end{center}
\vspace{-2em}
	\label{tabwflw}
\end{table*}
\subsection{Evaluation of Robustness against Occlusion}
Most state-of-the-art methods have got promising results on FLD under constrained environments. However, when the face image is suffering heavy occlusions and complicated illuminations, these methods will degrade greatly. In order to test the performance of our MMDN on faces with occlusions, we conduct the experiments on three heavy occluded benchmark datasets: the COFW dataset, the 300W challenging subset and the WFLW dataset.
\\\indent On the COFW dataset, the failure rate is defined by the percentage of test images with more than 10\% detection error. As illustrated in Table \ref{tab2}, we compare the proposed MMDN with other representative methods on the COFW dataset. Our MMDN boosts the NME to 5.01\% and the failure rate to 1.78\%, which outperforms the state-of-the-art methods. It suggests that our proposed IMCG model and EPBR method play an important role in boosting the ability to address the occlusion problems.
\\\indent We also compare our approach against the state-of-the-art methods on the 300W challenging subset in Table \ref{tab1} and Fig. \ref{fig6}. As shown in Table \ref{tab1}, our method achieves 6.08\% NME on the 300W challenging subset, which outperforms state-of-the-art methods on occluded faces. Furthermore, the Cumulative Error Distribution(CED) curve in Fig. \ref{fig6} also depicts that our model achieves superior performance in comparison with other methods. 
\\\indent The WFLW dataset contains complicated occluded subsets such as the Illumination subset, the Make-Up Subset and the Occlusion Subset. As shown in Table \ref{tabwflw}, our MMDN outperforms other state-of-the-art FLD methods, which also demonstrates the effectiveness of the proposed MMDN.
\\\indent Hence, from the experimental results on the COFW dataset, the 300W challenging subset and the WFLW dataset, we can conclude that 1) the IMCG model can be used to enhance the geometric constraints and context information for faces with heavy occlusions. 2) by fusing the boundary heatmap regression and landmark heatmap regression via the proposed EPBR method, we can generate more effective boundary-adaptive landmark heatmaps, and more accurate landmarks can be predicted. 3) by integrating the proposed IMCG model and EPBR method into a Multi-order Multi-constraint Deep Network (MMDN), our method is more robust against occlusions.
\begin{table}
\caption{Comparisons with state-of-the-art methods on the AFLW dataset. The error (NME) is normalized by face size. (\% omitted)}
\vspace{-1em}
\begin{center}
\begin{tabular}{p{3.8cm}p{1.6cm}p{1.6cm}}
\hline
Method & full & frontal \\
\hline
SDM \cite{xiong2013supervised}&	4.05&	2.94\\
PCPR \cite{Burgosartizzu2013Robust} & 3.73 & 2.87   \\
ERT \cite{Kazemi2014OneMF}&	4.35&	2.75\\
LBF \cite{ren2014face} & 4.25 & 2.74   \\
CCL \cite{Zhu2016UnconstrainedFA}&	2.72&	2.17\\
DAC-CSR(CVPR17) \cite{Feng2017DynamicAC}&	2.27&	1.81\\
SAN \cite{Dong2018StyleAN}(CVPR18)&	1.91&	1.85\\
ODN \cite{Zhu2019RobustFL}(CVPR19)&	1.63&	1.38\\
\hline
SHN 	&	2.46&	1.92\\
SHN+FCC &	2.28 &	1.78\\
SHN+MCC &	2.17 &	1.69\\
SHN+MSC &   1.74 &	1.48\\
SHN+IMCG&	1.62 &	1.37\\
SHN+EPBR&	2.05 &	1.58\\
\textbf{SHN+IMCG+EPBR (MMDN)} &	\textbf{1.41}&	\textbf{1.21} \\
\hline
\end{tabular}
\end{center}
\vspace{-1em}
\label{tab3}
\end{table}
\vspace{-1em}
\subsection{Evaluation of Robustness against Large Poses}
\begin{figure}[t]
	\begin{center}
		\includegraphics[width=0.92\linewidth]{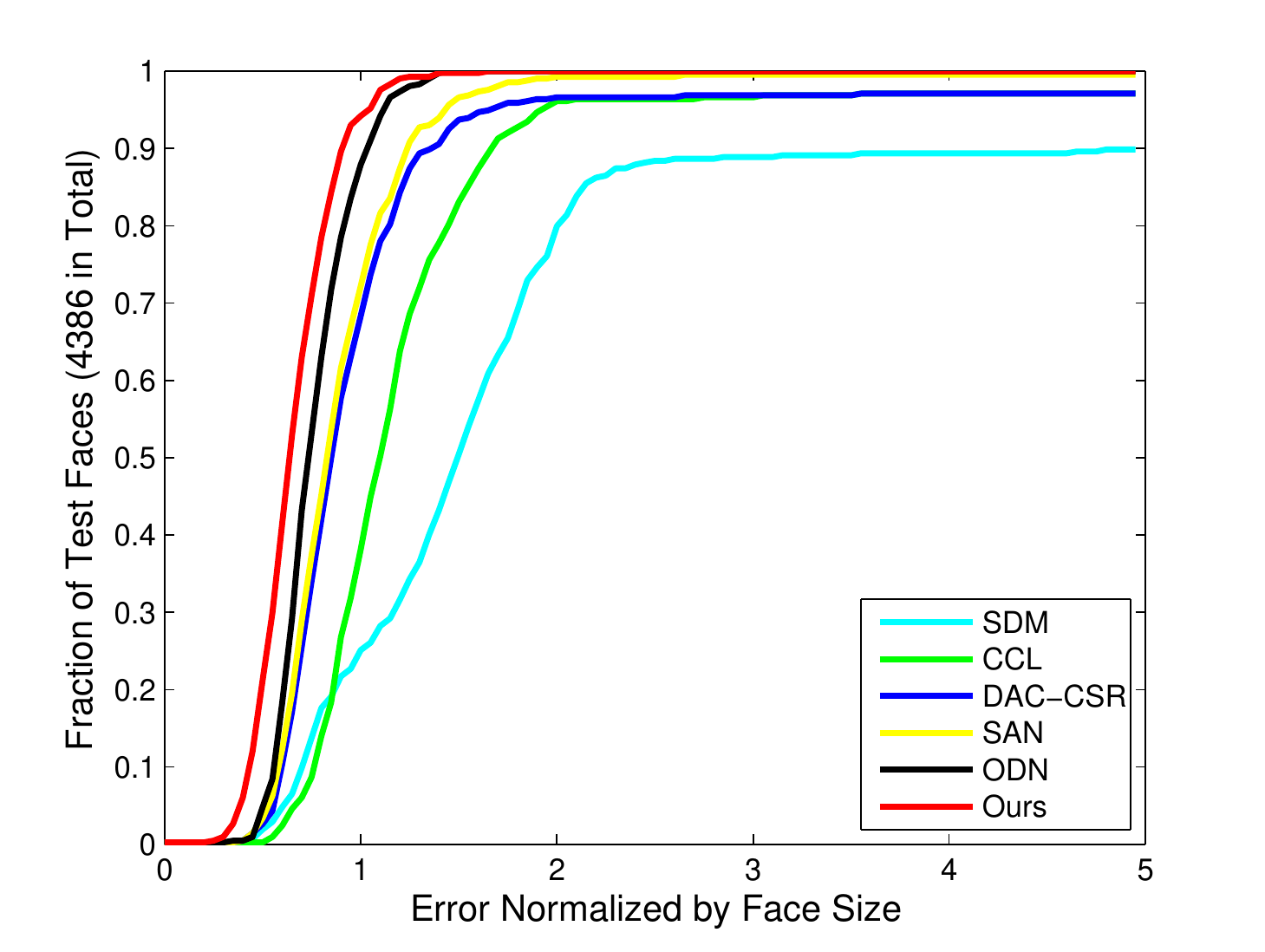}
	\end{center}
	\vspace{-1em}
	\caption{Comparisons of CED curves of our method and state-of-the-art methods like SDM \cite{xiong2013supervised}, CCL \cite{Zhu2016UnconstrainedFA}, DAC-CSR \cite{Feng2017DynamicAC}, SAN \cite{Dong2018StyleAN} and ODN \cite{Zhu2019RobustFL} on the AFLW-full dataset (19 landmarks). Our approach outperforms the other methods. Best viewed in color.}
	\vspace{-1em}
	\label{fig7}
\end{figure}

Faces with large poses are another great challenge for FLD. To further evaluate the effectiveness of our proposed method, we carry out experiments on the AFLW dataset, the 300W challenging subset and the WFLW dataset. For the AFLW dataset, Table \ref{tab3} shows that our method achieves 1.41\% NME on the AFLW-full testing set and 1.21\% NME on the AFLW-frontal testing set, which outperforms the state-of-the-art methods. Furthermore, the Cumulative Error Distribution(CED) curve in Fig. \ref{fig7} also depicts that our model exceeds the other methods. For the WFLW dataset, the NME on the Pose Subset and the Expression Subset surpasses the other methods. Hence, from the experimental results of the three datasets (see Table \ref{tab1}, \ref{tabwflw}, \ref{tab3}, Fig. \ref{fig6} and Fig. \ref{fig7}), we can conclude that our method is more robust to faces with large poses, mainly because the proposed IMCG model and EPBR method can utilize the implicit and explicit geometric constraints to generate more accurate boundary-adaptive landmark heatmaps and further improve the accuracy of FLD.
\subsection{Self Evaluations}
\textbf{Time and memory analysis.} The IMCG model contains multiple multi-order spatial correlations modules and one multi-order channel correlations module. The multi-order spatial correlations module only increases the computation costs as it introduces the matrix operations such as the matrix addition and the matrix outer product. The multi-order channel correlations module increases both the computation costs and memory costs, i.e., the first-order channel correlations block and the second-order channel correlations block introduce additional convolutional kernels $W_A$ and $W_C$. Moreover, the second-order channel correlations block also needs 0.125MB to store $Y_N$ and $Z_N$ for the $256 \times 256$ covariance matrix. The EPBR method only increases the computation costs that do not need to be calculated in the reference. During training, our MMDN takes thrice as long as SHN \cite{Yang2017StackedHN}, and in the reference, our MMDN takes twice as long as SHN that can reach 100FPS on a single tesla v100. Among all, compared to SHN, our MMDN introduces some additional parameters which are insignificant compared to 16GB memory on a tesla v100, and the impact of the consumed computation costs will be decreased with the rapid development of the hardware. To reduce the computation costs, we also conduct the experiment by replacing the last hourglass network unit of the SHN with the proposed MSC module, and the NME on the 300W challenging dataset will only rise from 6.63\% to 6.81\%, which still achieves state-of-the-art accuracy.
\begin{table}
	\renewcommand\arraystretch{1.2}
	\caption{The effect of loss functions on benchmark datasets (NME (\%))}
		\vspace{-1em}
	\begin{center}
		\begin{tabular}{p{1.4cm}|p{0.9cm}|p{0.7cm}p{0.7cm}p{0.7cm}p{0.7cm}p{0.7cm}}
			\hline
			\multirow{2}{*}{Datasets} & landmark & \multicolumn{5}{|c}{landmark and boundary} \\
			\cline{2-7}
			& {MSE1} & {MSE2} & {JSDL1} & {JSDL2} & {EPBR1} &{EPBR2} \\
			\cline{2-7}	
			\cline{0-0}
			{Challenging} & {7.56} & {7.34} & {7.10}  & {7.05} & {7.01} & {6.97} \\
			{COFW} & {6.21} & {6.04} & {5.82}  & {5.77} & {5.68} & {5.62} \\
			{AFLW-full} & {2.46} & {2.32} & {2.14} & {2.10} & {2.08} & {2.05} \\		 
			\hline
		\end{tabular}
		\vspace{-2em}
	\end{center}
	\label{tab4}
\end{table}
\\\indent \textbf{Loss function.} To evaluate the effectiveness of the proposed EPBR method, we conducted experiments on the 300W, COFW and AFLW datasets by separately using the Mean Squared Error (MSE1 and MSE2), the Jensen-Shannon Divergence Loss (JSDL1 and JSDL2) and our proposed EPBR (EPBR1, EPBR2) method. MSE1 means using the MSE to generate landmark heatmaps and traversing the generated landmark heatmaps to predict landmarks, i.e., the baseline SHN \cite{Yang2017StackedHN}. MSE2 and JSDL1 use MSE and JSDL separately to generate landmark heatmaps and boundary heatmaps, then they predict landmarks by traversing the generated landmark heatmaps. JSDL2 utilizes JSDL to generate landmark heatmaps and boundary heatmaps, and predicts landmarks by constructing and searching the boundary-adaptive landmark heatmaps. EPBR1 and EPBR2 both mean using the EPBR method to generate landmark heatmaps and boundary heatmaps, but EPBR1 predicts landmarks by traversing the generated landmark heatmaps, and EPBR2 predicts landmarks by searching the generated boundary-adaptive landmark heatmaps. From Table \ref{tab4}, we can conclude that: 1) generating landmark heatmaps and boundary heatmaps at the same time can improve the performance of FLD. 2) compared to MSE, the JSDL can help generate more effective landmark heatmaps and boundary heatmaps, which can help to detect more accurate facial landmarks. 3) the searching mechanism can not only be utilized to generate more accurate boundary-adaptive landmark heatmaps, but also can help to detect more accurate landmarks for the boundary-regression FLD methods as a post-processing method. 4) by integrating the JSDL and the searching mechanism via the proposed EPBR method, we can generate more robust boundary-adaptive landmark heatmaps, which can be further utilized to enhance the robustness to large poses and heavy occlusions.
\begin{figure*}
\begin{center}
\includegraphics[width=0.92\linewidth]{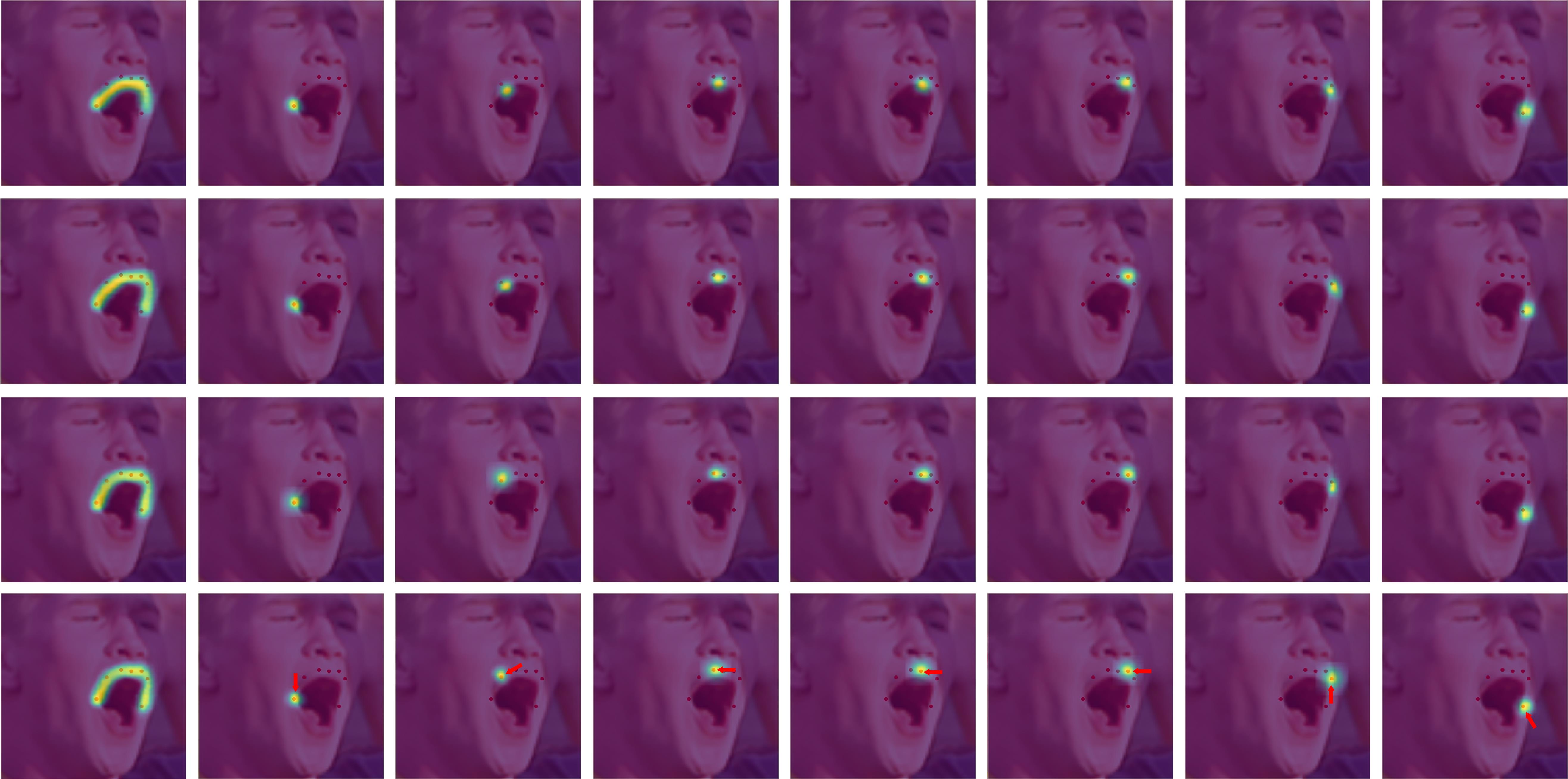}
\end{center}
\caption{Comparisons of predicted landmark heatmaps and boundary heatmaps corresponding to the upper side of the upper lip boundary. The first, second, third and fourth rows represent the boundary regression method with MSE, IMCG+MSE, IMCG+JSDL and IMCG+EPBR, respectively. Red dots represent the ground-truth landmarks. Our IMCG+EPBR can generate more accurate and consistent landmark heatmaps and boundary heatmaps, and the EPBR method can achieve landmark semantic moving (the third row shows the previous locations, red arrow represents the moving direction), which further help improve facial landmark detection by effectively utilizing the boundary information.}
\label{fig8}
\vspace{-2em}
\end{figure*}
\\\indent \textbf{Heatmap quality.} Comparisons of predicted landmark heatmaps and boundary heatmaps corresponding to the upper side of the upper lip boundary are shown in Fig. \ref{fig8}. The first, second, third and fourth rows represent the boundary regression method with MSE, IMCG+MSE, IMCG+JSDL and IMCG+EPBR, respectively. MSE means using the mean squared error (MSE) to generate landmark heatmaps and boundary heatmaps at the same time. IMCG+MSE, IMCG+JSDL and IMCG+EPBR mean combining the MSE, JSDL and our proposed EPBR method with the proposed IMCG model to generate landmark and boundary heatmaps, respectively. From Fig. \ref{fig8}, we have the following observations and corresponding analyses.
\\\indent (1) From the first row and the second row in Fig. \ref{fig8}, we can see that IMCG+MSE can generate more accurate landmark heatmaps and boundary heatmaps than MSE, mainly because our proposed IMCG model can effectively encode the geometric constraints and context information by introducing the multi-order spatial correlations and the multi-order channel correlations.
\\\indent (2) From the second row and the third row in Fig. \ref{fig8}, we can see that IMCG+JSDL outperforms IMCG+MSE. This indicates that generating heatmaps with the probability-based distribution constraints is more effective to enhance the accuracy and robustness of heatmap regression-based FLD than the pure pixel difference constraints, mainly because (a) the JSDL can accurately measure the difference between two distributions and pay more attention to the foreground area of heatmaps instead of equally treat the whole heatmap, (b) the probability-based regression method is closer to reality and with better universality and generalization.
\\\indent (3) From the third row and the fourth row in Fig. \ref{fig8}, we can see that IMCG+EPBR exceeds IMCG+JSDL. This indicates that (a) by integrating the JSDL and the searching mechanism via the proposed EPBR method, the quality of heatmaps can be further improved. (b) Compared to IMCG+JSDL, IMCG+EPBR can add semantically consistent constraints to the predicted landmarks by introducing the searching mechanism, i.e., the landmarks in facial boundary heatmaps can be semantically moved to improve the accuracy of FLD.
\subsection{Ablation Study}
Our proposed Multi-order Multi-constraint Deep Networks contain two pivotal components, namely, the IMCG model and the EPBR method. Hence, we conduct the following experiments.
\\\indent  (1) As the IMCG model contains the multi-order spatial correlations (MSC) module and multi-order channel correlations (MCC) module, we conduct experiments by combing the MSC module and the MCC module with the baseline SHN \cite{Yang2017StackedHN} respectively. As shown in Table \ref{tab1}, \ref{tab2} and \ref{tab3}, we can conclude that 1) performing feature re-calibration can help obtain more discriminative representations for robust FLD (SHN+FCC vs SHN). 2) the second-order statistics can help better model the channel correlations and detect more accurate landmarks (SHN+MCC vs SHN+FCC). 3) the multi-order spatial correlations can be utilized to enhance the geometrical constraints for robust FLD (SHN+MSC vs SHN). 4) By fusing the multi-order spatial correlations and the multi-order channel correlations, the accuracy of FLD can be further improved (SHN+IMCG vs (SHN+MSC and SHN+MCC)).
\\\indent (2) To evaluate the effectiveness of the proposed EPBR method, we conduct experiments by combing the SHN \cite{Yang2017StackedHN} with the EPBR method. The experimental results from benchmark datasets (see Table  \ref{tab1} -- \ref{tab4} and Fig. \ref{fig8}) demonstrate that by integrating the JSDL and the searching mechanism via the proposed EPBR method, we can explicitly add both boundary and semantically consistent constraints to the final predicted landmarks, achieving robust FLD.
\\\indent (3) Finally, the proposed MMDN is obtained by combining the SHN \cite{Yang2017StackedHN} with the IMCG model and the EPBR method. From the experimental results in Table \ref{tab1} -- \ref{tab3}, we can see that MMDN (SHN+IMCG+EPBR) surpasses the SHN+IMCG and the SHN+EPBR respectively, showing the complementary of these two components.
\section{Conclusion}
Unconstrained facial landmark detection is still a very challenging topic due to large poses and partial occlusions. In this work, we present a Multi-order Multi-constraint Deep Network (MMDN) to address FLD under extremely large poses and heavy occlusions. By fusing the Implicit Multi-order Correlating Geometry-aware (IMCG) model and the Explicit Probability-based Boundary-adaptive Regression (EPBR) method with a seamless formulation, our MMDN is able to achieve more robust FLD. It is shown that the IMCG model can enhance the representative and discriminative abilities of features and effectively encode the geometric constraints and context information by introducing the multi-order spatial correlations and multi-order channel correlations. The EPBR method can explicitly utilize the boundary information to enhance the geometric constraints and generate more effective boundary-adaptive landmark heatmaps, which further improve the performance of FLD. Experiments on benchmark datasets demonstrate that our method outperforms state-of-the-art FLD methods. It can also be found from the experiment that by fusing the searching mechanism and JSDL via the proposed EPBR method, the landmarks in a facial boundary can be semantically moved to further improve the accuracy of FLD.

%


\ifCLASSOPTIONcaptionsoff
  \newpage
\fi


%



\bibliographystyle{IEEEtran}
\bibliography{IEEEabrv,IEEEexample}
\end{document}